\documentclass[runningheads]{llncs}
\usepackage{graphicx}
\usepackage{tikz}
\usepackage{comment}
\usepackage{amsmath,amssymb} % define this before the line numbering.
\usepackage{color}

\usepackage{epsfig}
\usepackage{graphicx}
\usepackage{amsmath}
\usepackage{amssymb}
\usepackage{enumitem}
\usepackage[]{multirow}
\usepackage{booktabs}
\usepackage[accsupp]{axessibility}
\usepackage{mwe}
\usepackage{graphbox}
\usepackage{lipsum}
\usepackage{tabularx}
\usepackage{bm}
\usepackage{hyperref}
\hypersetup{pagebackref=true,breaklinks=true,letterpaper=true,colorlinks,bookmarks=false}
\usepackage[accsupp]{axessibility}  % Improves PDF readability for those with disabilities.
\begin{document}
% \renewcommand\thelinenumber{\color[rgb]{0.2,0.5,0.8}\normalfont\sffamily\scriptsize\arabic{linenumber}\color[rgb]{0,0,0}}
% \renewcommand\makeLineNumber {\hss\thelinenumber\ \hspace{6mm} \rlap{\hskip\textwidth\ \hspace{6.5mm}\thelinenumber}}
% \linenumbers
\pagestyle{headings}
\mainmatter
\def\ECCVSubNumber{7765}  % Insert your submission number here

\title{Super-resolution 3D Human Shape from a Single Low-Resolution Image} % Replace with your title implicit function representation of super-resolution shape reconstruction from Single Low-Resolution Images

% INITIAL SUBMISSION 
\begin{comment}
\titlerunning{ECCV-22 submission ID \ECCVSubNumber} 
\authorrunning{ECCV-22 submission ID \ECCVSubNumber} 
\author{Anonymous ECCV submission}
\institute{Paper ID \ECCVSubNumber}
\end{comment}
%******************

% CAMERA READY SUBMISSION
%\begin{comment}
\titlerunning{Super-resolution 3D Human Shape from a Single Low-Resolution Image}

% If the paper title is too long for the running head, you can set
% an abbreviated paper title here
%
\author{Marco Pesavento \and
Marco Volino \and
Adrian Hilton\inst{1}}%\orcidID{2222--3333-4444-5555}}
\authorrunning{M. Pesavento et al.}
% First names are abbreviated in the running head.
% If there are more than two authors, 'et al.' is used.
%
\institute{Centre for Vision, Speech and Signal Processing (CVSSP), University of Surrey, Guildford, UK 
\email{\{m.pesavento,m.volino,a.hilton\}@surrey.ac.uk}\\
}
%\end{comment}
%******************
\maketitle
\begin{abstract}
 We propose a novel framework to reconstruct super-resolution human shape from a single low-resolution input image. The approach overcomes limitations of existing approaches that reconstruct 3D human shape from a single image, which require high-resolution images together with auxiliary data such as surface normal or a parametric model to reconstruct high-detail shape. The proposed framework represents the reconstructed shape with a high-detail implicit function. Analogous to the objective of 2D image super-resolution, the approach learns the mapping from a low-resolution shape to its high-resolution counterpart and it is applied to reconstruct 3D shape detail from low-resolution images. The approach is trained end-to-end employing a novel loss function which estimates the information lost between a low and high-resolution representation of the same 3D surface shape. Evaluation for single image reconstruction of clothed people demonstrates that our method achieves high-detail surface reconstruction from low-resolution images without auxiliary data. Extensive experiments show that the proposed approach can estimate super-resolution human geometries with a significantly higher level of detail than that obtained with previous approaches when applied to low-resolution images. \href{https://marcopesavento.github.io/SuRS/}{https://marcopesavento.github.io/SuRS/}
% Quantitative evaluation against ground-truth 3D shape demonstrates that our method achieves improved reconstruction accuracy compared to previous single-image reconstruction methods that use only high resolution images without auxiliary data.
\end{abstract}
%%%%%%%%% BODY TEXT
\section{Introduction}
\begin{figure}[t!]
\parbox{.48\linewidth}{
\centering
\resizebox{1\linewidth}{!}{\begin{tabular}{cccc}
 1920x1080 & 256x256& PIFuHD\cite{saito2020pifuhd} & SuRS (Ours)\\ 
  \includegraphics[width=2.3in]{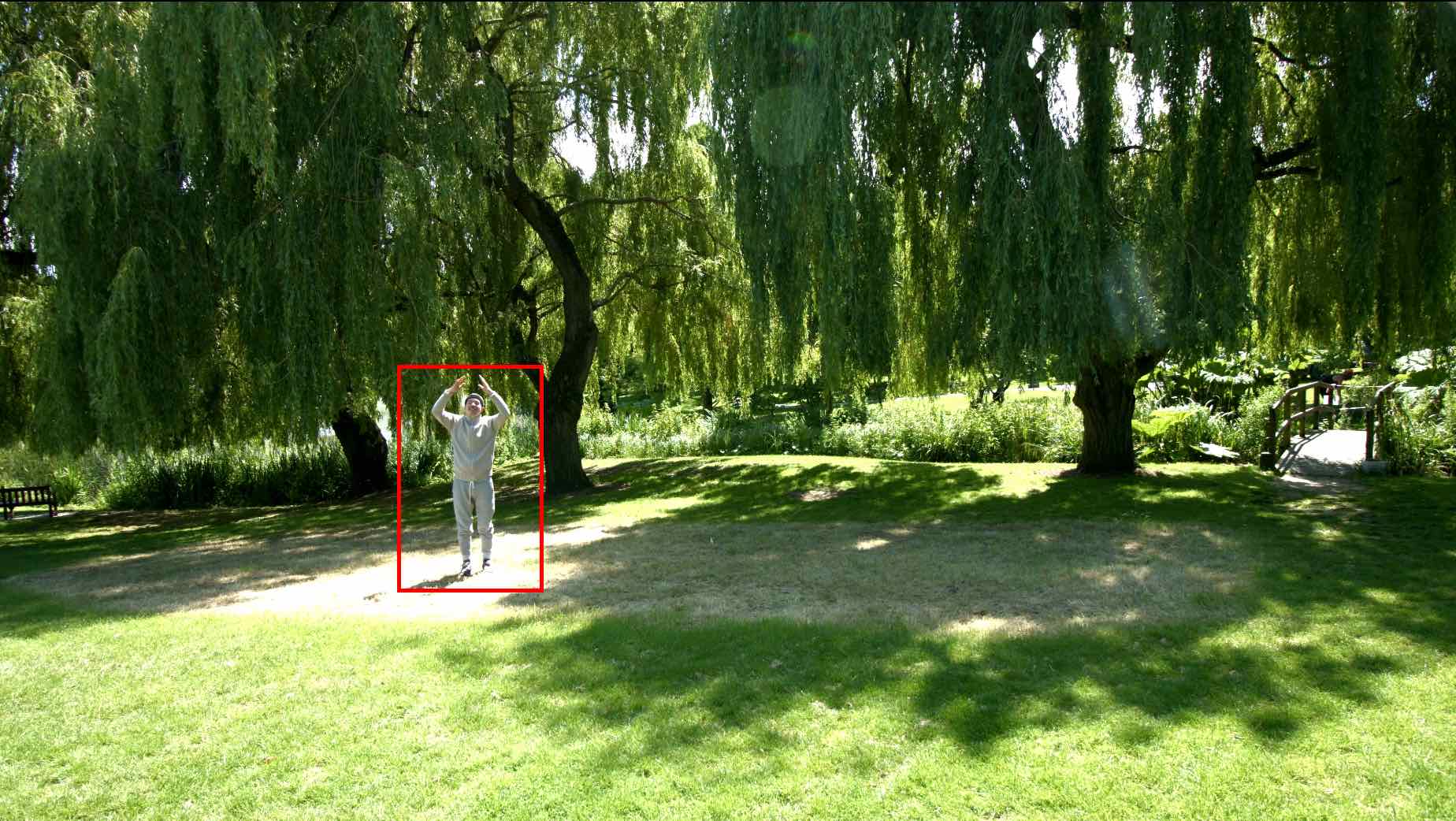}   &   \includegraphics[width=0.5in]{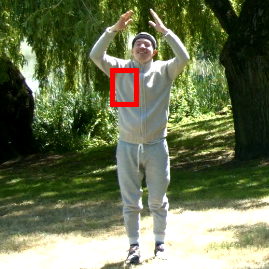}&\includegraphics[width=0.53in]{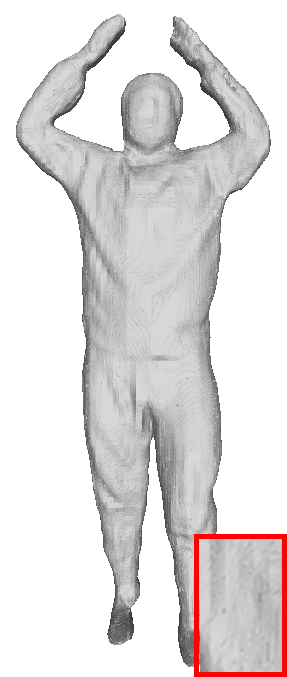}   &   \includegraphics[width=0.5in]{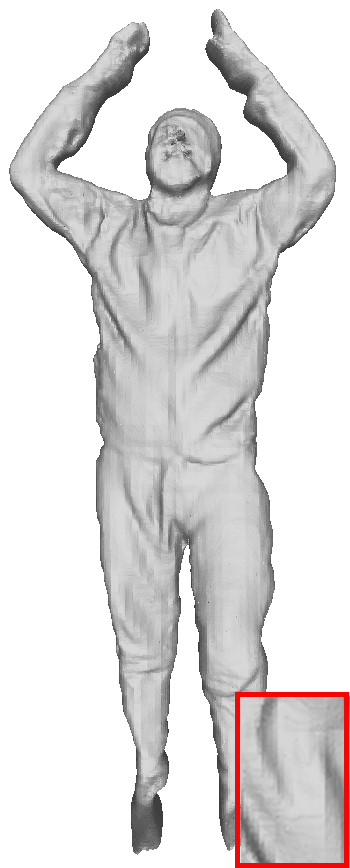} 
\end{tabular}}
 \caption{Person capture in a large space in a HR image~\cite{malleson2020real}. Our approach is able to represent significant higher level of fine details compared to PIFuHD~\cite{saito2020pifuhd} even if these are not clear in the LR input image.}
 %\vspace{-2.5em}
 \label{fig:introa}
}
\hfill
\parbox{.48\linewidth}{
\centering
\resizebox{0.85\linewidth}{!}{

\begin{large}

\begin{tabular}{ccccccc}
LR input             & \multicolumn{2}{c}{normal maps}     & \multicolumn{2}{c}{parametric model  }                  & only RGB image \\
image &  \multicolumn{2}{c}{PIFuHD~\cite{saito2020pifuhd}} & \multicolumn{2}{c}{PaMIR~\cite{zheng2021pamir}} & SuRS (ours) \\
\raisebox{0.4\height}{\includegraphics[width=0.5in]{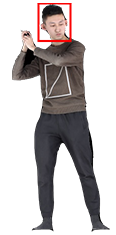}} & \includegraphics[width=0.5in]{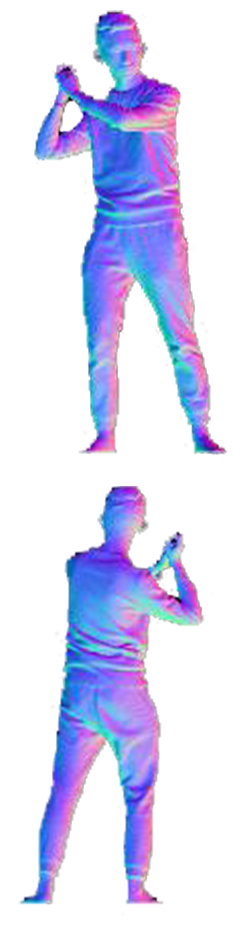} & \includegraphics[width=1.in]{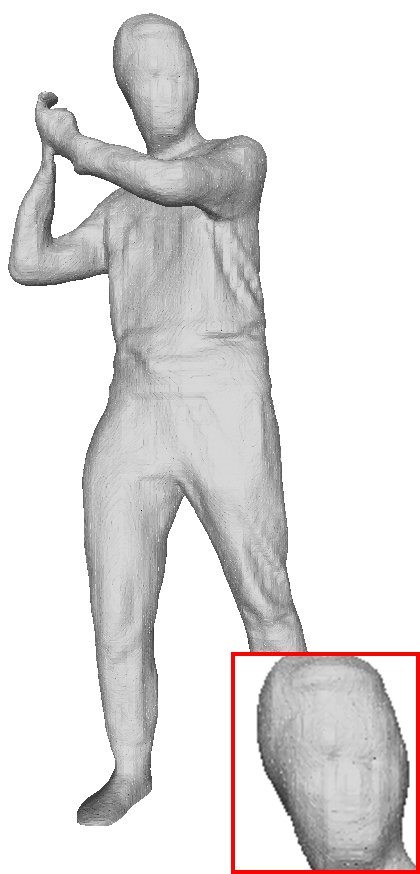} & \raisebox{0.3\height}{\includegraphics[width=0.5in]{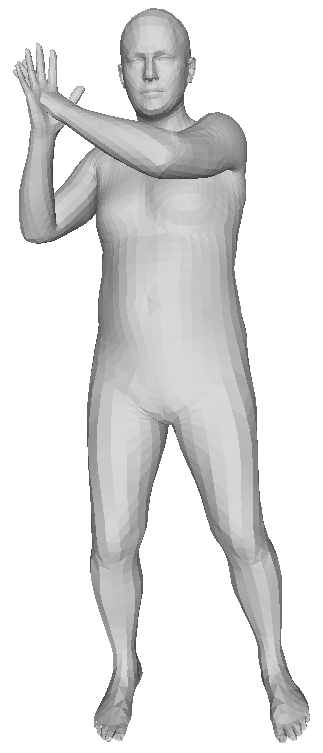}} & \includegraphics[width=1.in]{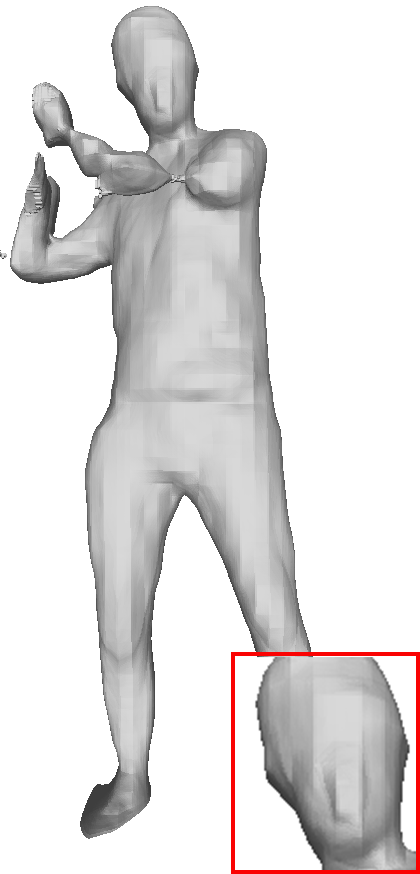} & \includegraphics[width=1.in]{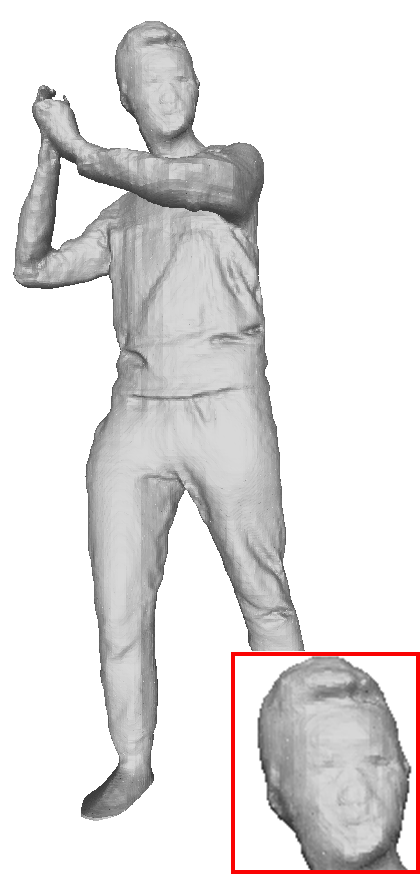}               
\end{tabular}
\end{large}

}
 %\vspace{-1.0em}
 \caption{3D human digitization of a low-resolution ($256\times256$) image. Our approach result represents significant higher level of details compared to related approaches that leverage auxiliary data.}
 %\vspace{-2.5em}
 \label{fig:introb}
}
\end{figure}

\noindent The demand for user-friendly frameworks to create virtual 3D content is increasing at the same pace as the rise in the application of immersive technologies. This has led research to a focus on finding practical solutions to replace existing sophisticated multi-view systems for 3D reconstruction, which are impractical and expensive to build for the general community. With the recent advance of 3D deep learning, several approaches have been proposed to estimate the 3D shape directly from a single RGB image for specific object classes such as human bodies, faces or man made objects. Although consumer cameras are nowadays able to capture high-resolution (HR) images, there are several scenarios where the image of a single person is not at the full camera image resolution but at a relatively low-resolution (LR) sub-image. Reconstruction of high quality shape from LR images of people is therefore important for example for images of multiple people, scenes requiring a large capture volume or when people are distant from the camera (Fig.~\ref{fig:introa}). Since the LR image contains little detail information, state-of-the-art approaches cannot represent fine details in the reconstructed shape. The introduction of pixel-aligned implicit function (PIFu~\cite{saito2019pifu}) improved the quality of human shapes reconstructed from a single image. However, it is highly dependent in the level of details represented on the input image and it fails to reconstruct HR shape from a LR image that does not contain information on fine details. Recent works leverage auxiliary information such as displacement or normal maps together with HR images ($1024\times1024$) to retrieve fine details~\cite{saito2020pifuhd,alldieck2019tex2shape}. Other methods combine the implicit function with 3D information by using voxel alignment and 3D convolution~\cite{he2020geo,zheng2021pamir,li2020detailed}. However, none of these works can reconstruct high quality shapes by only using a single LR RGB image.
\\To tackle this problem, we introduce a new framework that generates Super-Resolution Shape (SuRS) via a high-detail implicit function which learns the mapping from low to high-resolution shape. %The introduced concept of Super-Resolution for shapes indicates the high level of details presented in the reconstructed shapes retrieved by learning to map a low resolution shape $S_{LR}$ to its high resolution counterpart $S_{HR}$.
\label{sec:intro}
If a low-resolution shape $S_{LR}$ is compared with its high-resolution counterpart $S_{HR}$, the primary difference is in the fine details that represent the information lost between $S_{LR}$ and $S_{HR}$. This information is retrieved by learning the mapping from low to high-resolution shape such that corresponding fine details can be reproduced in the super-resolution shape $S_{SR}$. We apply SuRS to the task of 3D human digitization to estimate high-detail 3D human shape from a single LR RGB image ($256\times256$) without assisting the training with auxiliary data such as normal maps or parametric models. Our approach learns the missing information of $S_{LR}$ in order to estimate fine shape details from a LR image even if these details are not clearly visible in the input image. The final high-detail shape is represented with an implicit representation, which is learned by a novel end-to-end trainable framework with a new loss that estimates the information missing in the low-resolution shape $S_{LR}$ by computing the difference between the implicit representation of high-resolution shape $S_{HR}$ and the estimated implicit representation of the reconstructed super-resolution shape $S_{SR}$. This loss facilitates the learning of the map as shown in Section~\ref{sec:abl}. Low and high-resolution features are extracted from the LR input image and then given as input to two multi-layer perceptrons (MLPs), which estimate the probability field over the 3D space. The pixel-alignment feature used in previous approaches~\cite{saito2019pifu,saito2020pifuhd} is adopted by SuRS to preserve local details and infer plausible shape detail in unseen regions.
Extensive experiments show that SuRS reconstructs significantly higher resolution human shape detail compared to previous single image human reconstruction approaches when LR images are given as input (Fig.~\ref{fig:introa}). The level of details on the reconstructed surfaces is also much higher to that achieved by state-of-the-art approaches that leverage auxiliary data in the reconstruction process (Fig.~\ref{fig:introb}).
\\The main contributions of this work are:
\begin{itemize}[noitemsep]
    %\item A new concept for super-resolution shape represented by implicit function.
    \item An end-to-end trainable approach for implicit function learning of super-resolution shape for 3D human digitization from single low-resolution image.
    \item Introduction of a novel loss to learn the mapping between low and high resolution shapes based on the information loss between their representations.
    \item Improved quantitative and perceptual performance for human shape reconstruction from a single LR image compared with state-of-the-art approaches.
\end{itemize}
%%%%%%%%%%%%%%%%%%%%%%%%%%%%%%%%%%%%%%%%%
\section{Related Works}
%\subsection{3D super-resolution}
\indent \textbf{3D super resolution: }The concept of super resolution (SR) is well established in the 2D domain. On the other hand, there is considerably less work related to super resolution applied to the 3D domain. Most approaches focus on super-resolving 2.5D data such as texture maps~\cite{li20193d,richard2019learned,pesavento2021attention,pesavento2021super}, depth maps~\cite{voynov2019perceptual,ni2017color,song2020channel,sang2020inferring}, or light field images~\cite{rossi2018geometry,zhang2021end}. All of these methods apply similar network architectures to the ones used to super-resolve 2D images. They consider the 2.5D data as a RGB image and leverage a modified convolutional neural network (CNN) architecture with additional data such as normal maps or edge images as input. Sinha et al.~\cite{sinha2017surfnet} and Chen et al.~\cite{chen2018synthesizing} first retrieve a geometry image that embeds geometric information and then super-resolve it with a standard SR CNN, reconstructing the final HR mesh. In this case, the super-resolution task consists of creating the HR mesh by subdividing the LR mesh through remeshing and super-resolving the extracted geometry images. In other methods~\cite{wu2019point,dinesh2020super}, the super-resolution task is formulated for point cloud data as increasing the number of points in the final model, achieved by applying graph networks. 
Li et al.~\cite{li2020detailed} first retrieve a coarse estimation of human shape from multi-view images and then refine the reconstruction by voxelizing the shape and applying 3D convolution to super-resolve the voxel. The SR voxel is obtained by subdividing each voxel grid in order to increase their total number in the final voxel. We introduce a new concept for super-resolution of shape represented via high-detail implicit function. In this paper, super resolution is defined as the reproduction of high quality details in the reconstructed shape even if they are not embedded in the input. To overcome the resolution limitation of previous methods, we propose a novel approach that learns to map a LR shape to its HR counterpart.
%%%%%%%%%%%%%%%%%%%%%%%%%%%%%%%%%%%%%%%%%%%%%%%iinsert.   implicit function work
%\subsection{3D Human digitization}
\\\indent \textbf{3D Human digitization:} Early 3D human digitization strategies such as fitting the input image by a parametric model~\cite{bogo2016keep,kanazawa2018end,pavlakos2019expressive,kocabas2020vibe,xu20213d}, using a 3D CNN to regress volumetric representation~\cite{varol2018bodynet,zheng2019deephuman} or adding a displacement on each vertex of the mesh~\cite{alldieck2018video,alldieck2019tex2shape,alldieck2019learning}, mainly aim to recover the naked body's shape with only approximate representation of shape details such as clothing and hairs. With the ground-breaking introduction of implicit functions for 3D reconstruction~\cite{mescheder2019occupancy,park2019deepsdf,chen2019learning,huang2020arch,he2021arch++}, deep learning approaches that estimate 3D human shape, started to adopt this form of representation because it can represent fine-scale details on the reconstructed surfaces. The first work to adopt the implicit representation for 3D human digitization from a single image was PIFu~\cite{saito2019pifu}, which exploits pixel-aligned image features rather than global features. The local details present in the input image are preserved since the occupancy of any 3D point is predicted. 
\\ There are two main problems with this approach that are addressed from other works: it may generate incorrect body structures and struggle to recover fine-scale geometric details. With regards to the first problem, several frameworks combine the implicit function with either a parametric model~\cite{huang2020arch,zheng2021pamir} or geometry-aligned shape feature~\cite{he2020geo} to improve the stability of prediction since they serve as a shape prior for the reconstruction. These works focus on improving the parts of the reconstructed geometry that are not seen in the input image, deliberately neglecting the representation of fine details. To increase the quality of the reconstructed shape, PIFuHD~\cite{saito2020pifuhd} leverages HR images ($> 1k$ resolution) as well as front and back normal maps in the reconstruction process. The normal maps need to be predicted from the HR input image with an additional network. The use of HR images and normal maps increases the computation cost and memory use. Other approaches reconstruct high-detail shape using multiple images \cite{hong2021stereopifu,zhang2021dimnet,li2020detailed,zins2021data}, which is a different objective from single image 3D reconstruction.
\\Our proposed framework aims to reconstruct super-resolution geometries from a single LR image ($256\times256$) without leveraging auxiliary data.
%-------------------------------------------------------------------------
\section{Methodology}
\noindent We introduce a novel framework that aims to learn a high-detail implicit shape representation to recover Super-Resolution Shape (SuRS) from a single LR RGB image. A LR image does not contain significant information about fine details and previous approaches for shape reconstruction from a single image cannot reconstruct detailed shape if the fine detail are not embedded in the input (Section~\ref{sec:comp}). 
%is it right to say mapping implicit representation?
Inspired by the 2D image super-resolution objective of mapping LR images to their HR counterparts, SuRS learns to map a LR surface $S_{LR}$ to its HR counterpart $S_{HR}$. By learning the missing information of $S_{LR}$ with respect to $S_{HR}$, our approach can reproduce the fine details of the HR shape even if they are not embedded in the input. We define Super-Resolution shape $S_{SR}$ as a shape with fine details reconstructed from a LR input image.% that contains little information about the fine details.
%------------------------------------------------------------------
\subsection{Implicit Function Representations of Super-Resolution Shape}
\begin{figure}[t!]
    \centering
    \resizebox{0.85\linewidth}{!}{
    \includegraphics[width=1.in]{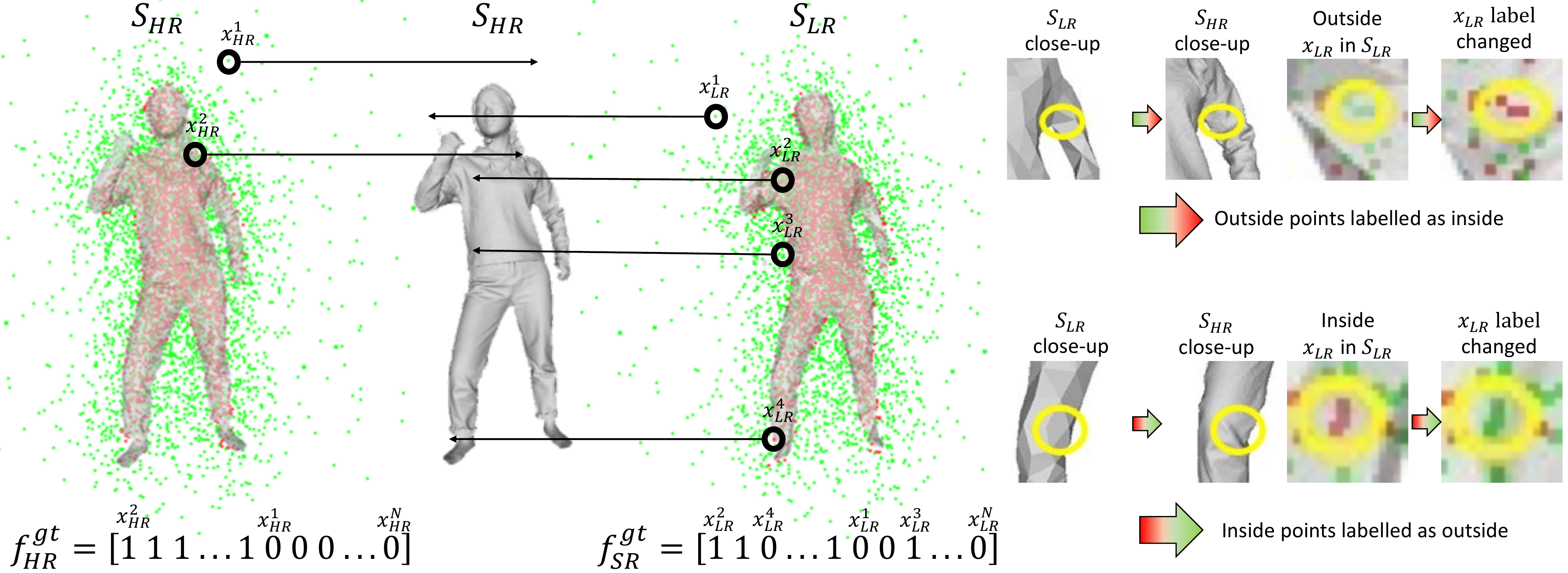}}
    %\vspace{-2.1em}
    \caption{On the left, comparison between ordinary implicit representation of 3D points $x_{HR}$ sampled from the HR surface $S_{HR}$ space and implicit function representation of super-resolution shape with 3D points $x_{LR}$ sampled from the LR surface $S_{LR}$ space and classified with respect to $S_{HR}$. The red points are `inside' the surfaces while the green ones are `outside'. On the right, close-up of the labelling process: some points `outside' $S_{LR}$ are represented as `inside' points (first row) while some other points `inside' $S_{LR}$ are labelled as `outside' (second row).} % In $l=1$, similarity mapping is performed between input subvectors and every part of every reference, in $l=2$ between input subvectors and all the parts of a single reference, in $l=3$ between input subvectors and all the references.}
    %%\vspace{-1.8em}
    \label{fig:sampling}
\end{figure}
\noindent An implicit function defines a surface as a level set of a function $f$, e.g. $f(X) = 0$~\cite{sclaroff1991generalized} where $X$ is a set of 3D points in $\Re^3$. To represent a 3D surface $S$, this function $f(X)$ is modelled with a Multi-Layer Perceptron (MLP) that classifies the 3D points as either `inside' or `outside' the 3D surface $S$. A HR surface $S_{HR}$ can be represented as a 0.5 level-set of a continuous 3D occupancy field:
\begin{align}
f_{HR}^{gt}(x_{HR}) =  \begin{cases} 1, \text{if $x_{HR}$ is inside $S_{HR}$} \\ 0, \text{otherwise} \end{cases}
\end{align}
where $x_{HR}$ is a point in the 3D space around the HR surface $S_{HR}$.
This means that $f_{HR}^{gt}$ is a vector of length $N$, where $N$ is the total number of 3D points $x_{HR}$ that are sampled from $S_{HR}$ space and classified as `inside' or `outside' points:\\
$\bullet f_{HR}^{gt}[0:\frac{N}{2}]=1$ correspond to the $\frac{N}{2}$ 3D points inside $S_{HR}$; \\
$\bullet f_{HR}^{gt}[\frac{N}{2}+1:N]=0$ correspond to the $\frac{N}{2}$ 3D points outside $S_{HR}$;
\\
To map the LR surface to the HR surface, we adapt the implicit function representation to the 3D super-resolution shape. We define a new ground truth for the high-detail implicit function of the estimated super-resolution shape $S_{SR}$:
\begin{align}
f^{gt}_{SR}(x_{LR}) =  \begin{cases} 1, \text{if $x_{LR}$ is inside $S_{HR}$} \\ 0, \text{otherwise} \end{cases}
\end{align}
where $x_{LR}$ are the 3D points sampled from the space of the LR surface $S_{LR}$. In contrast to the common implicit function, $f^{gt}_{SR}$ is created by classifying the 3D points $x_{LR}$ with respect to the HR surface instead of the LR one: some $x_{LR}$ points are labelled as `outside' even if they are `inside' $S_{LR}$, and viceversa (Fig.~\ref{fig:sampling}). The vector $f^{gt}_{SR}$ contains $N_{inside}\ne\frac{N}{2}$ values equal to 1 and $N_{outside}\neq N_{inside}\ne\frac{N}{2}$ values equal to 0.
This will lead the MLP that models $f^{gt}_{SR}$ to classify points of the LR surface $S_{LR}$ as points of an HR surface $S_{HR}$, learning a map from $S_{LR}$ to $S_{HR}$. The points that change classification represent the difference between $S_{HR}$ and $S_{LR}$, which is primarily in the fine details of the shapes. SuRS learns to infer this difference to estimate super-resolution shape detail from a LR image. %the estimated shape creating high quality details even without significant information embedded on the input.
\subsection{Super-Resolution Single-View Shape Estimation}
\begin{figure*}[t!]
    \centering
    \resizebox{0.9\linewidth}{!}{
    \includegraphics[width=1in]{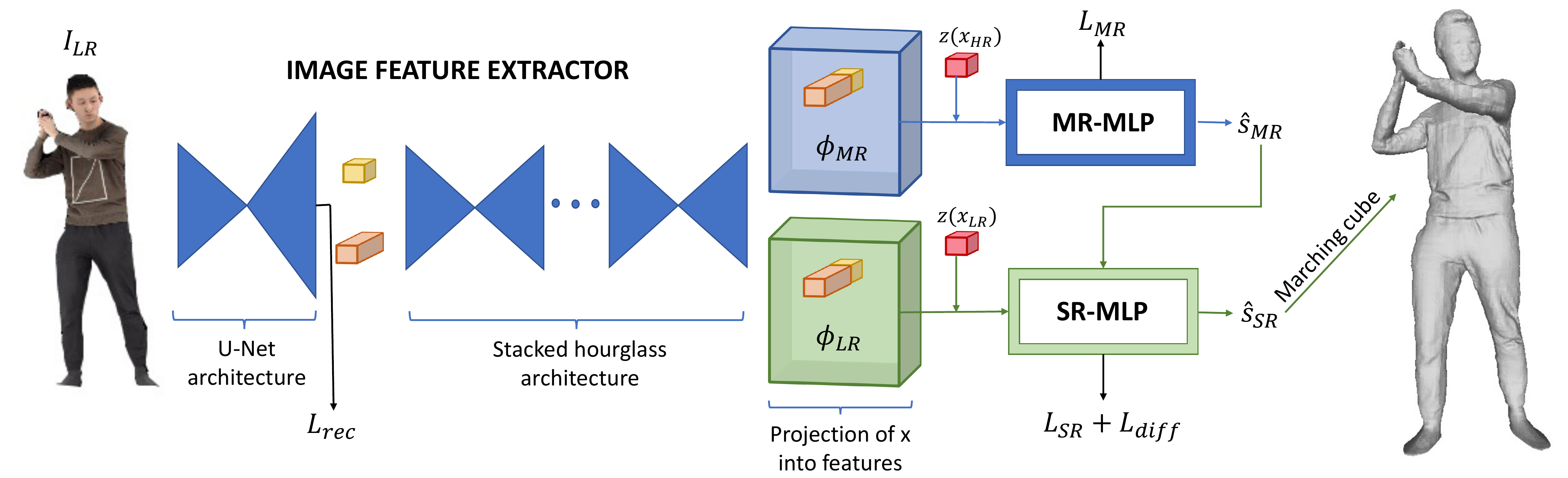}}
    %\vspace{-1.em}
    \caption{Network architecture of SuRS. The image feature extractor creates an embedding from the LR input image. 3D points are projected on the embedding and classified first by MR-MLP to produce a mid-level resolution estimation and then by SR-MLP to obtain the super-resolution estimation, from which the 3D SR shape is reconstructed.} % In $l=1$, similarity mapping is performed between input subvectors and every part of every reference, in $l=2$ between input subvectors and all the parts of a single reference, in $l=3$ between input subvectors and all the references.}
    %\vspace{-2em}
    \label{fig:arch}
\end{figure*}
\noindent We apply SuRS to the task of 3D human digitization from a single LR RGB image. To reconstruct a 3D human shape from a single high-resolution RGB image $I_{HR}$, Saito et al.~\cite{saito2019pifu} introduced pixel aligned implicit function (PIFu), whose objective is to predict the binary occupancy value for any given 3D position $x$ in continuous camera space. Instead of classifying directly the 3D points $X$ of the surface, PIFu estimates the level-set by exploiting the image feature $\phi(p)$ of the 2D projection $p=\pi(x_{HR})$ of the 3D point $x_{HR}$ in the input image $I_{HR}$. The function $f$ is represented by a MLP that is conditioned on the depth value $z(x_{HR})$ of $x_{HR}$ in the camera coordinate space:
\begin{equation}
\label{eq:eq4}
    f_{HR}(\phi(p,I_{HR}),z(x_{HR}))=\hat{s}_{HR}, \hat{s}_{HR} \in \mathbb{R}
\end{equation}
We modify the pixel-aligned implicit function representation by adapting it to the reconstruction of super-resolution shape from a low-resolution image $I_{LR}$:
\begin{equation}
\label{eq:eq2}
    f_{SR}(\phi(p_{LR},I_{LR}),z(x_{LR}),\hat{s}_{MR})=\hat{s}_{SR}, \hat{s}_{SR} \in \mathbb{R}
\end{equation}
where $x_{LR}$ are the 3D points sampled from the space of the LR surface $S_{LR}$, $p_{LR}=\pi(x_{LR})$, $\hat{s}_{MR}$ is a mid-resolution estimation of the shape computed with $f_{HR}^{gt}(x_{HR})$ from a LR input image and $\hat{s}_{SR}$ is a super-resolution estimation of the shape computed with $f^{gt}_{SR}(x_{LR})$.
SuRS is modelled via a neural network architecture composed of 3 modules and trained end-to-end (Fig.~\ref{fig:arch}). 
\\\textbf{$\bullet$ Image features extractor: }an image feature embedding is extracted from the projected 2D location of the 3D points in the LR image $I_{LR}$. Specifically, the embedding is the concatenation of two feature vectors extracted from the input image of size ($N_{I} \times N_{I}$): one with the resolution of ($\frac{N_{I}}{2} \times \frac{N_{I}}{2}$) to maintain holistic reasoning and the second of higher resolution ($2N_{I} \times 2N_{I}$) to embed the fine detail of the input image. The combination of holistic reasoning and high-resolution images features is essential for high-fidelity 3D reconstruction~\cite{saito2020pifuhd}. 
\\
\textbf{$\bullet$ Mid-resolution multi-layer perceptron (MR-MLP): }a first classification $\hat{s}_{MR}$ of the 3D points $x_{HR}$ sampled from $S_{HR}$ space is computed with an MLP that models the pixel aligned implicit function $f_{MR}(\phi(p,I_{LR}),z(x_{HR}))$. 
$x_{HR}$ are projected through orthogonal projection to the LR input image $I_{LR}$ and indexed to the feature embedding retrieved before. This is concatenated with the depth value $z(x_{HR})$ of $x_{HR}$ in the camera coordinate space and given as input to MR-MLP. $f_{MR}$ differs from $f_{HR}$ because the input is a low-resolution image: the output representation of MR-MLP does not reconstruct high-detail shape since information is missing in the LR input image and it cannot be learned with the standard implicit function modelled by MR-MLP (Section~\ref{sec:abl}).
\\
\textbf{$\bullet$ Super-resolution multi-layer perceptron (SR-MLP): }the final estimation $\hat{s}_{SR}$ is obtained with a further MLP, which models the implicit function $f_{SR}$ to represent super-resolution shape. %that models $f_{SR}$ of equation \ref{eq:eq2}. 
The input of SR-MLP differs from the MR-MLP one since the 3D points $x_{LR}$ are sampled from the space of the LR surface $S_{LR}$, projected to the input image and indexed to the feature embedding. Their projection is then concatenated with the depth value $z(x_{LR})$ of $x_{LR}$. To help the learning of a map from $S_{LR}$ to $S_{HR}$, the estimation $\hat{s}_{MR}$ is concatenated with the input embedding of SR-MLP to facilitate the learning of the lost information of $S_{LR}$. SR-MLP trained with $f^{gt}_{SR}$ learns to classify $x_{LR}$ according to the difference between $S_{LR}$ and $S_{HR}$. Compared to MR-MLP, SR-MLP additionally infers this difference to the estimation $\hat{s}_{SR}$, representing fine details on the SR shape even if they are not represented in the LR input image.
\\During inference, a LR image is given as input to the network. A set of $M$ random 3D points $x$ of the 3D space are projected on the extracted embedding and then classified first with MR-MLP to estimate $\hat{s}_{MR}$ and then with SR-MLP to estimate $\hat{s}_{SR}$.
The final shape is obtained by extracting iso-surface $f=0.5$ of the probability field $\hat{s}_{SR}$ at threshold 0.5 applying the Marching Cube algorithm~\cite{lorensen1987marching}.
\subsection{Training Objectives}
\noindent Our objective function combines 4 different losses used to train the 3 modules:
\begin{equation}
    \mathcal{L}=\mathcal{L}_{rec}+\mathcal{L}_{MR}+\mathcal{L}_{SR}+\mathcal{L}_{diff}
\end{equation}
\textbf{2D reconstruction loss: }the higher resolution feature extracted from the LR input image is double the size of the image. To ensure that the newly created features will preserve the spatial structure of the LR input image, SuRS minimises the L1 loss commonly used in the 2D image super-resolution task~\cite{wang2020deep}:
\begin{equation}
    \mathcal{L}_{rec}=||I_{SR}-I_{GT}||_1
\end{equation}
$I_{SR}$ is an image reconstructed from the HR features extracted by the feature extractor module and processed by a further convolutional layer. $I_{GT}$ is the ground-truth of the reconstructed image $I_{SR}$. %write on implementation how it is retrieved
\\\textbf{MR loss:}
since MR-MLP represents a pixel aligned implicit function, we train the network by minimising the average of mean squared error between MR-MLP output and the canonical ground-truth of implicit functions as done in PIFu~\cite{saito2019pifu}:
\begin{equation}
    \mathcal{L}_{MR}=\frac{1}{N}\sum_{i=1}^{N}|f_{MR}(\phi^i_{MR},z(x^i_{HR}))-f_{HR}^{gt}(x^i_{HR})|^2
\end{equation}
where $N$ is the number of points $x_{HR}$ sampled from the space of $S_{HR}$, $\phi^i_{MR}=\phi(p,I_{LR})$ and $f_{MR}(\phi^i_{MR},z(x^i_{HR}))=\hat{s}_{MR}$.
\\\textbf{SR loss:}
to compute the final estimation, the average of mean squared error between the output of SR-MLP and SuRS ground-truth is minimised:
\begin{equation}
\scalebox{0.9}{
    $\mathcal{L}_{SR}=\frac{1}{N}\sum_{i=1}^{N}|f_{SR}(\phi^i_{LR},z(x^i_{LR}),\hat{s}_{MR})-f_{SR}^{gt}(x^i_{LR})|^2$}
\end{equation}
where $\phi^i_{LR}=\phi(p_{LR},I_{LR})$, $f_{SR}(\phi^i_{LR},z(x^i_{LR}),\hat{s}_{MR})=\hat{s}_{SR}$ and the $N$ points $x_{LR}$ are sampled from the space of $S_{LR}$.
\\\textbf{Difference loss: }to facilitate the learning of the map from $S_{LR}$ to $S_{HR}$, we design a novel loss that minimises the average of mean squared error of the difference between $S_{SR}$ and $S_{HR}$, expressed as the difference between the estimated values $\hat{s}_{MR}$ and $\hat{s}_{SR}$:
\begin{equation}
    \mathcal{L}_{diff}=\frac{1}{N}\sum_{i=1}^{N}|(f^{gt}_{HR}-f^{gt}_{SR})-(\hat{s}_{MR}-\hat{s}_{SR})|^2
\end{equation}
Since $f^{gt}_{SR}$ contains the missing information of $S_{LR}$, this loss leads the network to learn in which points $S_{LR}$ differs from $S_{HR}$. Since these points represent the fine details, the network learns how to reproduce them in the estimated shape. This loss is important to learn the super-resolution shape detail (see Section~\ref{sec:abl}).
\section{Experiments}
\noindent We quantitatively and qualitatively evaluate the proposed approach on the task of reconstructing 3D human shape from a single LR image. To train SuRS, LR surfaces must be retrieved from a set of HR example models. The performance will improve if the displacement between $S_{HR}$ and $S_{LR}$ is higher as shown in Section~\ref{sec:abl}. Therefore, we use the THuman2.0~\cite{tao2021function4d} dataset, which consists of $524$ high-resolution surfaces which have a high level of fine details. We split the dataset into training (402 models) and testing (122 models). To evaluate the generalisation of the method, we test SuRS on 136 models taken from 3D People~\cite{3Dpeople}. 
We evaluate the reconstruction accuracy with 3 quantitative metrics: the normal reprojection error introduced in~\cite{saito2019pifu}, the average point-to-surface Euclidean distance (P2S) and the Chamfer distance (CD), expressed in cm.
\subsection{Implementation Details}
\noindent SuRS is trained with LR images $I_{LR}$ of size ($N_{I} \times N_{I}, N_{I}=256$) obtained by downscaling by a factor of 2 ground truth images $I_{GT}$ with bicubic degradation. We first apply~\cite{Barill:FW:2018} to make the HR surfaces watertight. For training, a LR shape $S_{LR}$ is obtained by remeshing $S_{HR}$ from $\sim400k$ to $1k$ faces via quadric edge collapse decimation~\cite{garland1998simplifying}. $N_T=24000$ 3D points $X$ are sampled with a mixture of uniform sampling and importance sampling around surface $S_{HR}$ following PIFu~\cite{saito2019pifu}. A subset of $X$ of $N=6000$ $x_{HR}$ points are sampled for $S_{HR}$ by selecting $\frac{N}{2}$ points inside and $\frac{N}{2}$ outside $S_{HR}$. A different subset of $X$ of $N=6000$ $x_{LR}$ points are sampled for $S_{LR}$ by selecting $\frac{N}{2}$ points inside and $\frac{N}{2}$ outside $S_{LR}$. The novel architecture of the image feature extractor is a combination of U-Net~\cite{ronneberger2015u} and stacked hourglass architectures~\cite{newell2016stacked}. The former retrieves the features whose resolution is double that of the input image while the second has been proved to be effective for surface reconstruction~\cite{newell2016stacked}. We design a novel U-Net architecture to retrieve both high and low-resolution features from the input image (see supplementary material). The HR features are then processed by a convolution layer while the LR features by 3 stacks. The outputs of the image feature extractor are a HR feature vector of size ($2N_{I} \times 2N_{I} \times 64$) and a LR one of size ($\frac{N_{I}}{2} \times \frac{N_{I}}{2} \times 256$). The final embedding is obtained with the concatenation of these two vectors. MR-MLP has the number of neurons of (321, 1024, 512, 256, 128, 1) while SR-MLP of (322, 1024, 512, 256, 128, 1) with skip connections at $3^{rd}$, $4^{th}$ and $5^{th}$ layers. 
\subsection{Ablation Studies}
\label{sec:abl}
\begin{figure}[t!]
\parbox{.48\linewidth}{
\centering
\resizebox{0.95\linewidth}{!}{\begin{Huge}
\begin{tabular}{ccccc}
LR input image             &1 &2&3               & 4 (ours) \\
\includegraphics[width=3.2in]{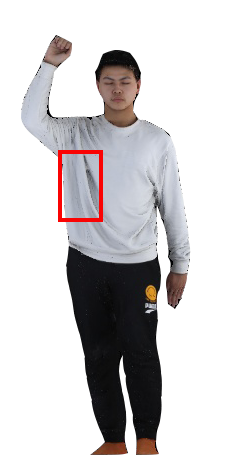}  & \includegraphics[width=2.55in]{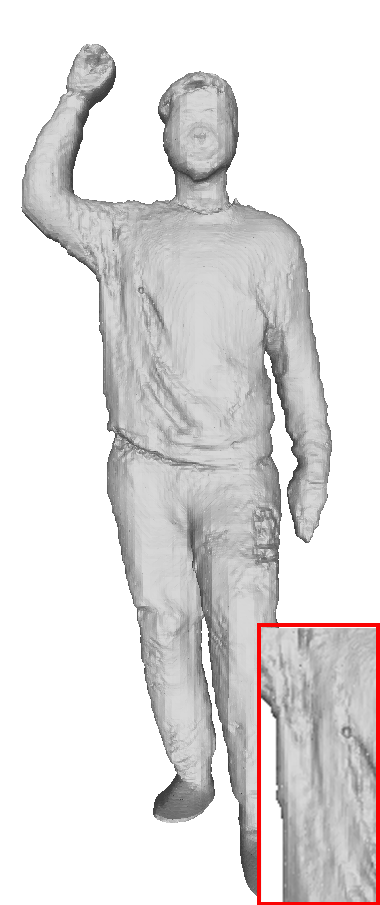} & \includegraphics[width=2.4in]{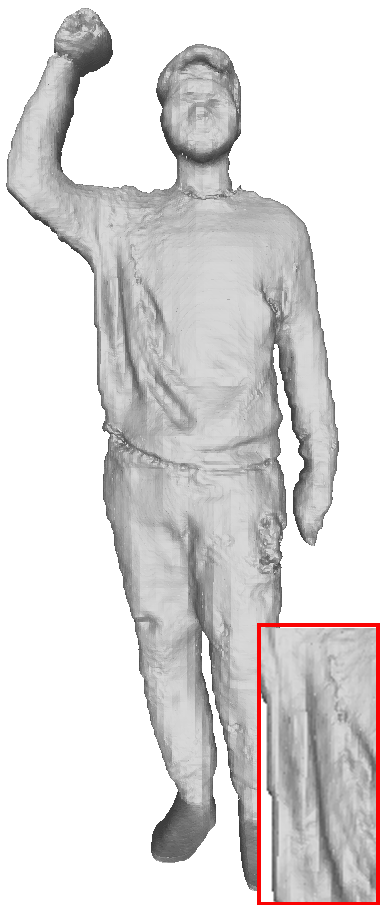} & \includegraphics[width=2.5in]{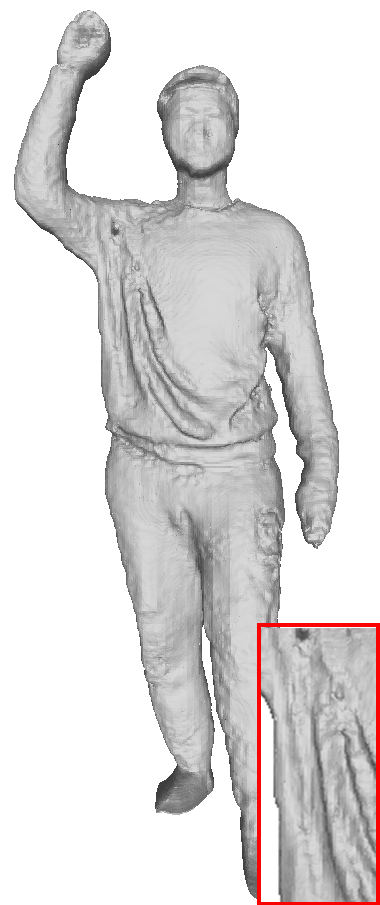} & \includegraphics[width=2.5in]{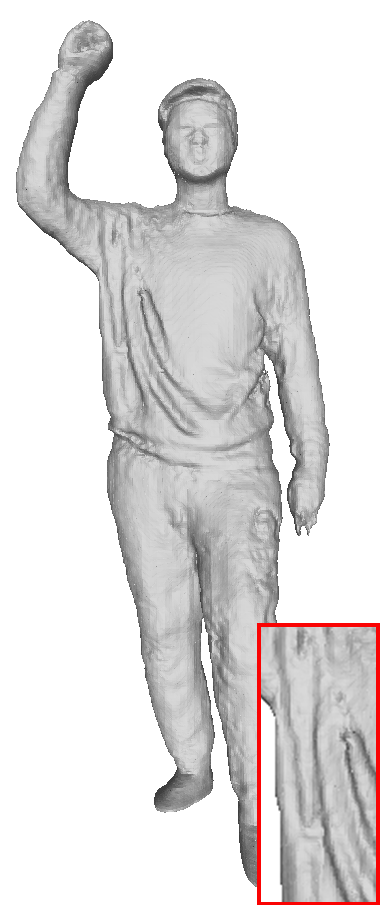}     \\   
\midrule
\includegraphics[width=3.in]{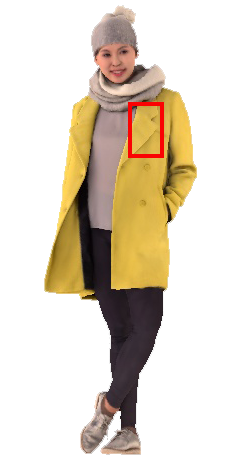}  & \includegraphics[width=2.in]{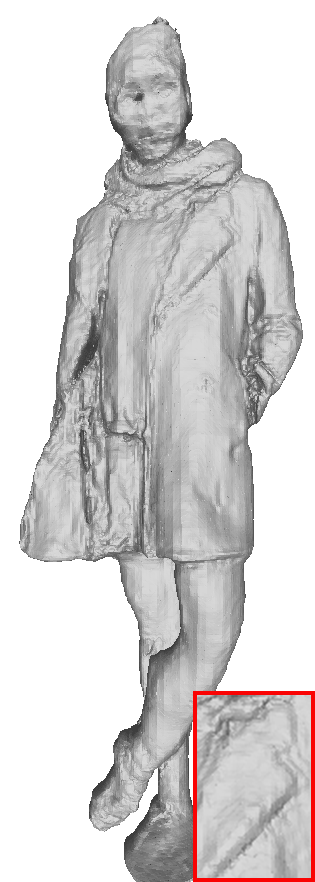} & \includegraphics[width=2.in]{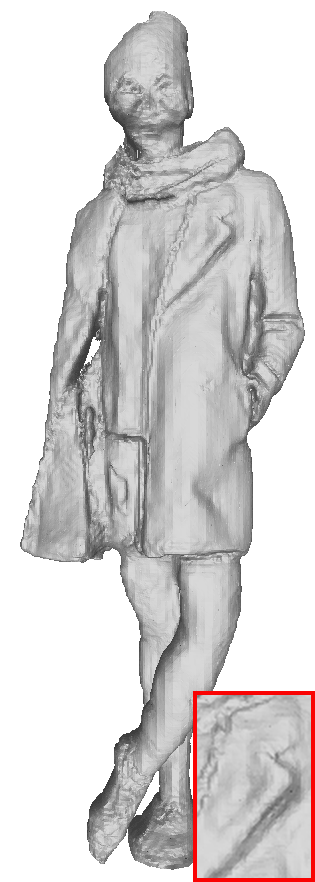} & \includegraphics[width=2.in]{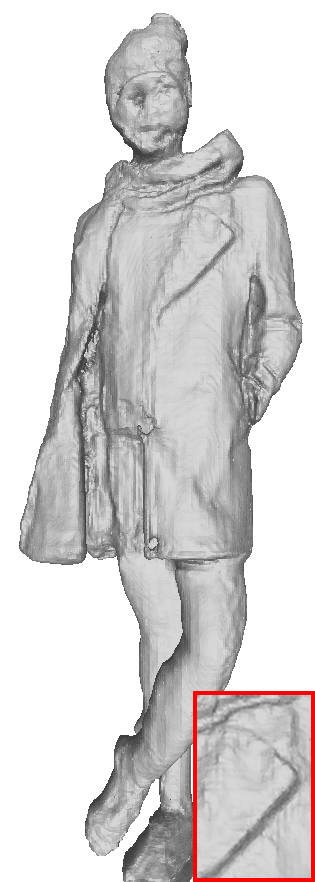} & \includegraphics[width=2.in]{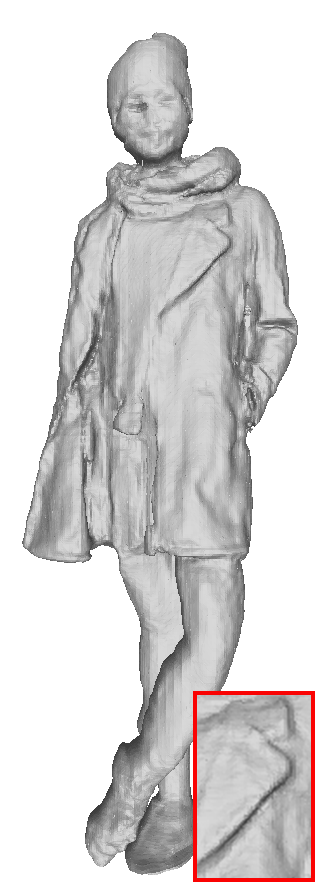} 
\end{tabular}
\end{Huge}}
%\vspace{-0.8em}
 \caption{Visual results obtained by changing the configuration of training. 1, 2, 3 and 4 are different configurations (refer to Table~\ref{tbl:train}). The upper model is from THuman2.0, the below one from 3DPeople.}
 %\vspace{-1em}
 \label{fig:train}
}
\hfill
\parbox{.48\linewidth}{
    \centering
    \resizebox{1\linewidth}{!}{\begin{Huge}
\begin{tabular}{ccccc}
LR input image             &10000 faces  &50000 faces&1000000 faces                  & 1000 faces (ours) \\
\includegraphics[width=2.3in]{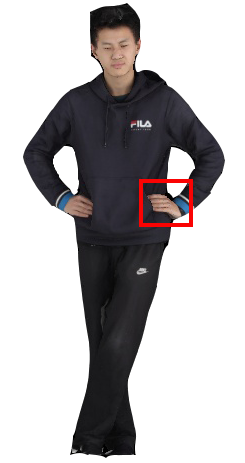}  & \includegraphics[width=2in]{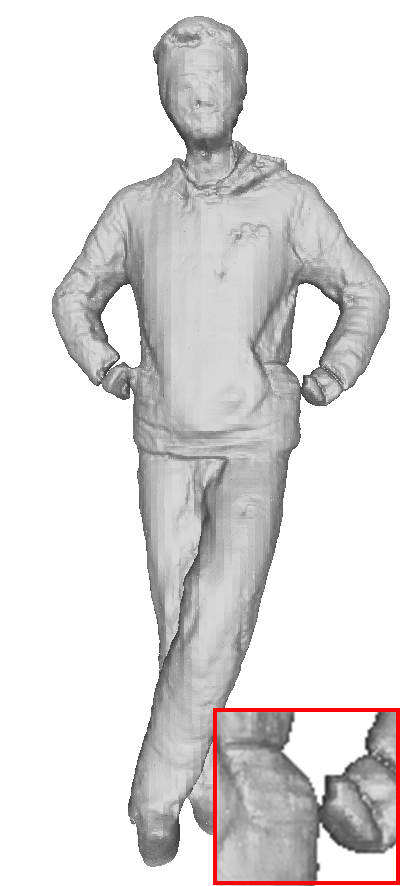} & \includegraphics[width=2in]{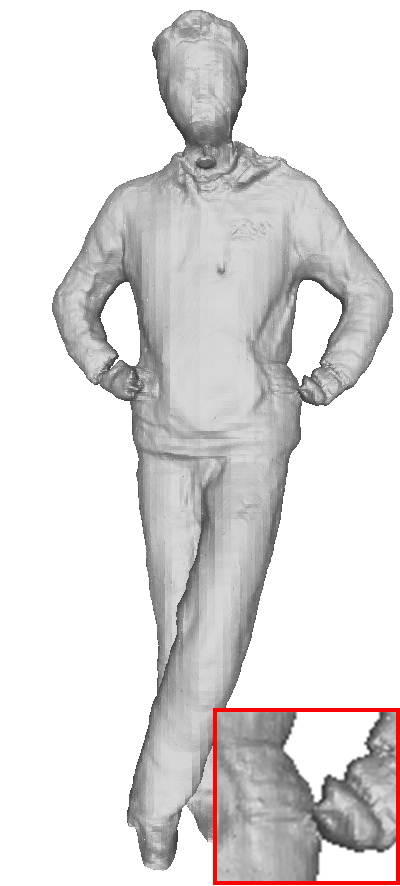} & \includegraphics[width=2in]{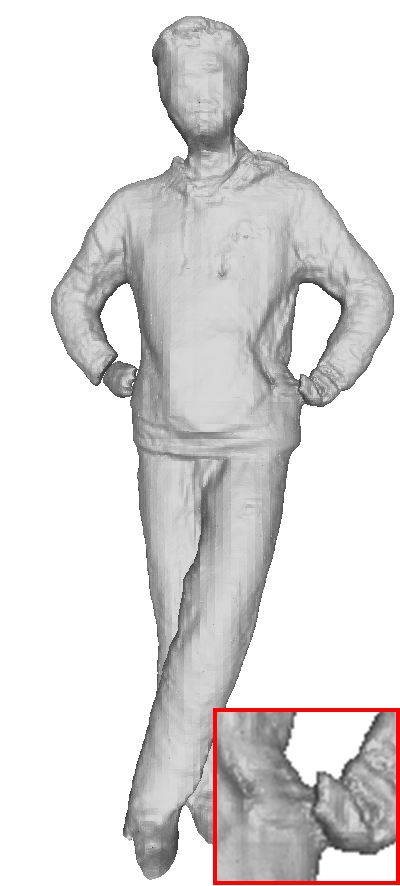} & \includegraphics[width=2in]{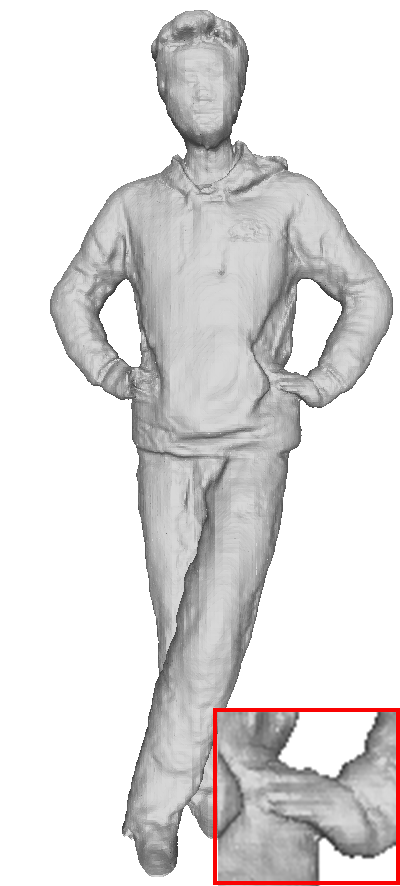}\\
\midrule
\includegraphics[width=2.5in]{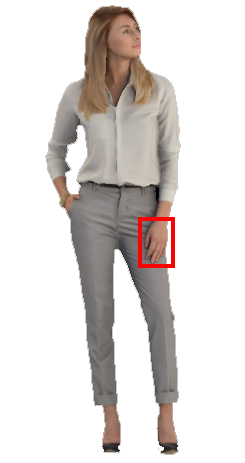}  & \includegraphics[width=2in]{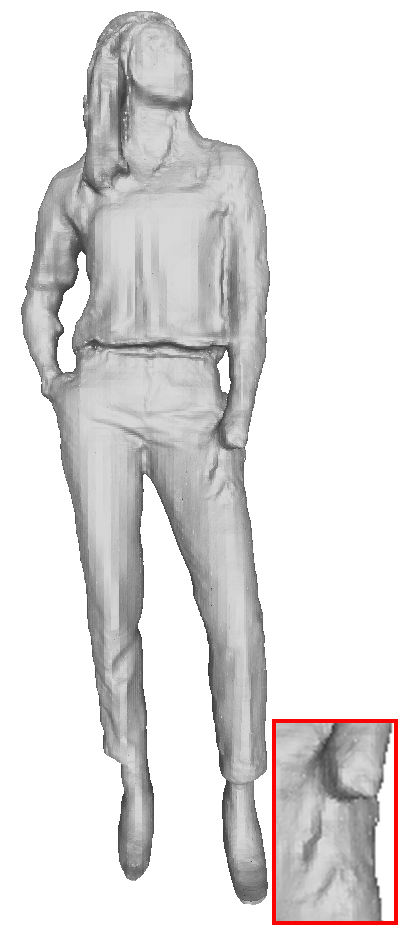} & \includegraphics[width=2in]{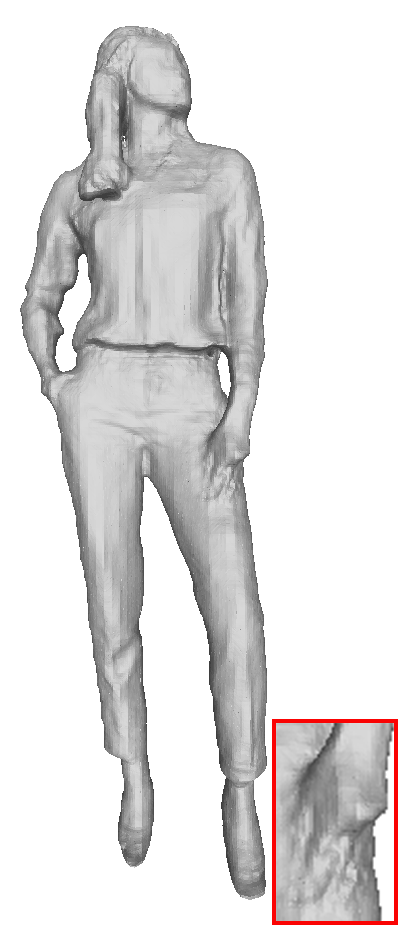} & \includegraphics[width=2in]{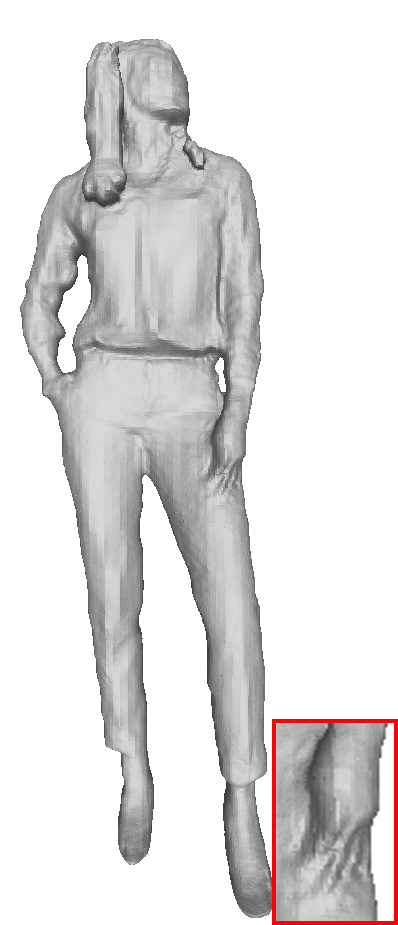} & \includegraphics[width=2in]{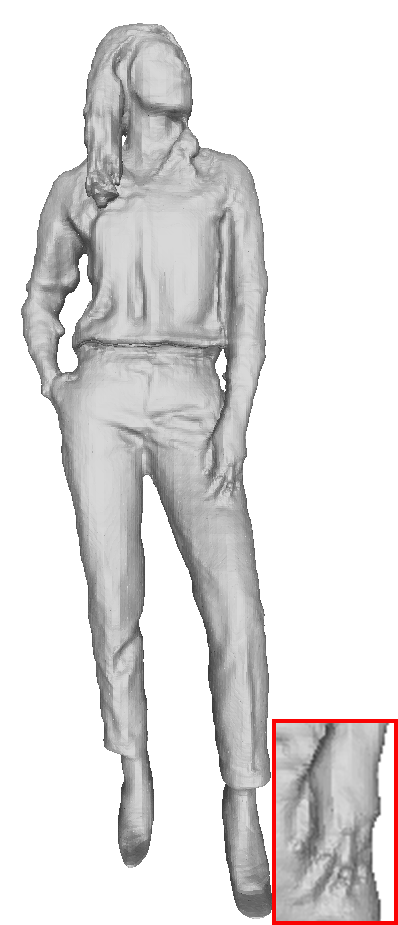} 
\end{tabular}
\end{Huge}}
%\vspace{-1.0em}
 \caption{Qualitative results obtained by using different decimation factors to create the low-resolution ground-truth shape. The upper model is from THuman2.0, the below one from 3DPeople.}
%\vspace{-2.3em}
 \label{fig:deg}
}
\end{figure}

\indent\textbf{Training configurations. }In the first ablation study we demonstrate that training the losses of the network in an end-to-end manner improves the performance of SuRS. We evaluate the network by training separately the three modules; by training first the feature extractor ($\mathcal{L}_{rec}$) and then the two MLPs together ($\mathcal{L}_{diff} + \mathcal{L}_{MR} + \mathcal{L}_{SR}$); by first training the feature extractor and the MR-MLP together ($\mathcal{L}_{rec} + \mathcal{L}_{MR}$) and then the SR-MLP ($\mathcal{L}_{diff} + \mathcal{L}_{SR}$). $\mathcal{L}_{diff}$ is always trained with $\mathcal{L}_{SR}$ since it depends on the output of SR-MLP. The values of the considered metrics are the highest when the end-to-end training is adopted (Table~\ref{tbl:train}). This is confirmed by the qualitative evaluation (Fig.\ref{fig:train}). The worst results are obtained when the feature extractor ($\mathcal{L}_{rec}$) is trained separately from the MLPs (case 1, 2). In these cases, the feature extractor learns how to create the embedding without being influenced by the classification part of the network: the extracted features embed information that is meaningless for the task of 3D human reconstruction. When the feature extractor is trained along with the MLPs (case 3, 4), the extracted features are influenced by the other losses and they are more informative for the objective of 3D human digitization. This proves that just applying a 2D image super-resolution network to upscale the input image as a pre-processing step is less efficient than the proposed approach. Training SR-MLP separately introduces noise in the reconstructed shapes while the end-to-end configuration produces the best results.
\\\indent\textbf{Decimation factors. }To create the LR geometry from the HR shape, we apply quadric edge collapse decimation that reduces the number of faces. We evaluate our approach by changing the decimation factor of the number of faces of $S_{LR}$. Low-resolution geometries with 100000, 50000, 10000 and 1000 faces are created (examples are illustrated in the supplementary material). Both the quantitative (Table~\ref{tbl:deg}) and qualitative (Fig.~\ref{fig:deg}) results show that the performance increases when the number of faces of $S_{LR}$ is the lowest, with significantly sharper details on the output shapes. Our approach is more efficient if the difference between the HR ground-truth shape and the LR ground-truth one is higher. SuRS learns to estimate missing information to obtain the SR shape by mapping $S_{LR}$ to $S_{HR}$. As $S_{LR}$ is coarser, the information loss increases while if $S_{LR}$ is similar to $S_{HR}$, there is no missing information, hence nothing to learn. The LR shapes with 1000 faces are the ones with the highest displacement from their HR counterparts, hence the improvement in the performance.
\begin{table}[t!]
\parbox{.48\linewidth}{
    \centering
    \caption{Quantitative results obtained by training SuRS with different configurations.}
\resizebox{0.5\textwidth}{!}{\begin{tabular}{|c|lll|lll|}
\hline
\multirow{2}{*}{\textbf{\begin{tabular}[c]{@{}c@{}}Training\\ Configuration\end{tabular}}} &
\multicolumn{3}{c|}{\textbf{THuman2.0}}                   & \multicolumn{3}{c|}{\textbf{3D people}} \\ \cline{2-7} 
                                                                                           & CD                              & Normal & P2S            & CD       & Normal             & P2S     \\ \hline
1: $\mathcal{L}_{rec}$;$\mathcal{L}_{MR}$;$\mathcal{L}_{SR}+\mathcal{L}_{disp}$               & 1.375                           & 0.1360 & 1.454          & 1.509    & 0.1261             & 1.636   \\
2: $\mathcal{L}_{rec}$;$\mathcal{L}_{MR}+\mathcal{L}_{SR}+\mathcal{L}_{disp}$                 & 1.261                           & 0.1347 & 1.496          & 1.243    & 0.1240             & 1.470   \\
3: $\mathcal{L}_{rec}+\mathcal{L}_{MR}$;$\mathcal{L}_{SR}+\mathcal{L}_{disp}$                 & 1.035                           & 0.1106 & \textbf{1.083} & 1.121    & 0.1160             & 1.281   \\
4: End to end (ours)                                                                          & \textbf{0.931} & \textbf{0.1065} & 1.151          & \textbf{1.057}    & \textbf{0.1127}    & \textbf{1.247}  \\ \hline
\end{tabular}}
%\vspace{-0.8em}
%\vspace{-2em}
\label{tbl:train}
}
\hfill
\parbox{.48\linewidth}{
\centering
\caption{Quantitative evaluation of using different decimation factors to create the low-resolution ground-truth shape.}
\resizebox{0.5\textwidth}{!}{\begin{tabular}{|l|lll|lll|}
\hline
\multirow{2}{*}{\textbf{Nr. faces}} & \multicolumn{3}{c|}{\textbf{THuman2.0}}                            & \multicolumn{3}{c|}{\textbf{3D people}}           \\ \cline{2-7} 
                                    & CD                              & Normal          & P2S            & CD             & Normal          & P2S            \\ \hline
100000                              & 0.942                           & 0.1084          & 1.117          & 1.147          & 0.1123          & 1.331          \\
50000                               & 0.950                           & 0.1079          & 1.117          & 1.118          & \textbf{0.1120}         & 1.291          \\
10000                               & 0.941                           & 0.1113          & \textbf{1.116} & 1.084          & 0.1133          & 1.276          \\
1000 (ours)                         & \textbf{0.931} & \textbf{0.1065} & 1.151          & \textbf{1.057} & 0.1127 & \textbf{1.247} \\ \hline
\end{tabular}}
%\vspace{-0.8em}
%\vspace{-2.2em}
\label{tbl:deg}
 }
\end{table}
\\\indent\textbf{Efficiency of the proposed architecture and displacement loss. }
We demonstrate the significant improvement achieved by the combination of the 3 modules and by the introduction of $\mathcal{L}_{diff}$. We train and test 4 frameworks obtained by modifying the architecture of SuRS:
\begin{itemize}[noitemsep]
\item\textbf{Without U-Net:} the U-Net architecture of the feature extractor is not implemented. The feature embedding, extracted only by the stacked hourglass part of the future extractor, is composed by two feature vectors of size ($N_{I} \times N_{I} \times 64$) and ($\frac{N_{I}}{2} \times \frac{N_{I}}{2} \times 256$). $\mathcal{L}_{rec}$ is not minimised during training.
\item\textbf{Only MR-MLP:} this framework consists only of the image feature extractor and MR-MLP. SR-MLP is not implemented and the shape is reconstructed from $\hat{s}_{MR}$. $\mathcal{L}_{SR}$ and $\mathcal{L}_{diff}$ are not considered during training.
\item\textbf{Only SR-MLP:} the feature embedding is directly processed by SR-MLP that outputs the estimation $\hat{s}_{SR}$ without being conditioned on $\hat{s}_{MR}$. $\mathcal{L}_{MR}$ and $\mathcal{L}_{diff}$ are not minimised during training.
\item\textbf{Without $\mathcal{L}_{diff}$: }this framework is the same as the proposed one but it is trained without considering $\mathcal{L}_{diff}$ to check the importance of this loss.
\end{itemize}
\begin{table}[t!]
    \centering
    \caption{Quantitative results obtained by changing the architecture of the approach.}
\resizebox{0.5\textwidth}{!}{\begin{tabular}{|l|lll|lll|}
\hline
\multirow{2}{*}{\textbf{Architecture}} & \multicolumn{3}{c|}{\textbf{THuman2.0}}                                             & \multicolumn{3}{c|}{\textbf{3D people}}           \\ \cline{2-7} 
                                       & CD                              & Normal          & P2S                             & CD             & Normal          & P2S            \\ \hline
W/o U-Net                        & 1.090                           & 0.1228          & 1.204                           & 1.260          & 0.1296          & 1.364          \\
Only MR-MLP                            & 1.044                           & 0.1066          & 1.077                           & 1.240          & 0.1129          & 1.276          \\
Only SR-MLP                            & 1.039                           & 0.1146          & \textbf{1.051} & 1.223          & 0.1215          & 1.255          \\
W/o $\mathcal{L}_{diff}$               & 1.020                           & 0.1138          & 1.196                           & 1.127          & 0.1200          & 1.255          \\
SuRS (ours)                      & \textbf{0.931} & \textbf{0.1065} & 1.151                           & \textbf{1.057} & \textbf{0.1127} & \textbf{1.247} \\ \hline
\end{tabular}}
%\vspace{-0.8em}
%\vspace{-2.3em}
\label{tbl:loss}
\end{table}
\begin{figure*}[t!]
\centering
\resizebox{0.85\linewidth}{!}{\begin{Huge}
\begin{tabular}{cccccc}
LR input image             & W/o $U-Net$ &Only MR-MLP & Only SR-MLP  & W/o $\mathcal{L}_{diff}$            & SuRS (ours) \\
\includegraphics[width=3.5in]{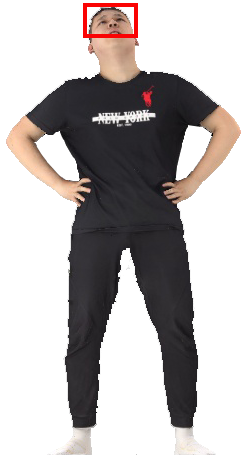}  & \includegraphics[width=4in]{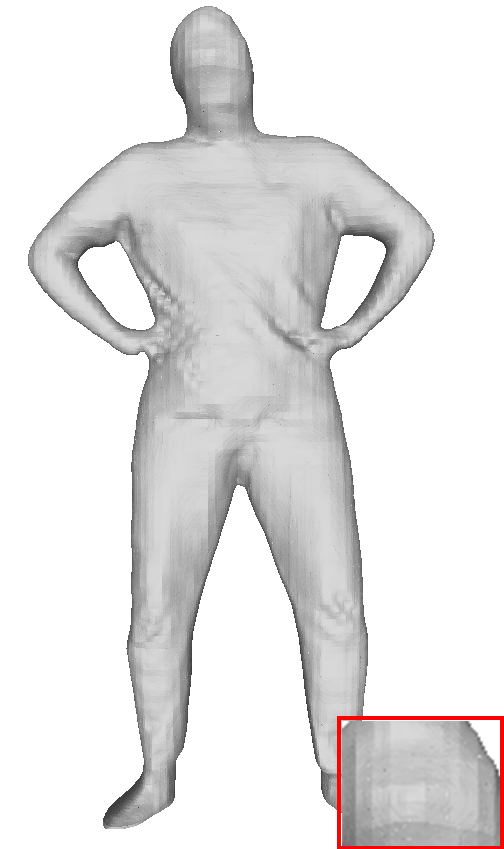} & \includegraphics[width=4in]{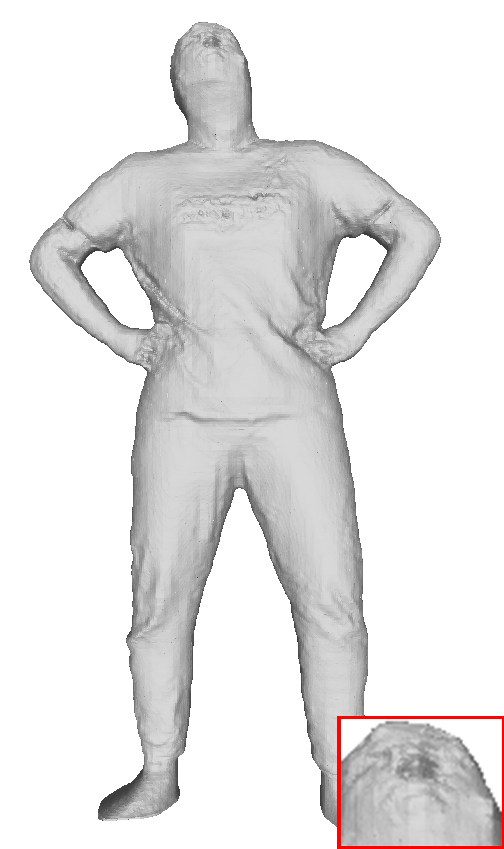} & \includegraphics[width=4in]{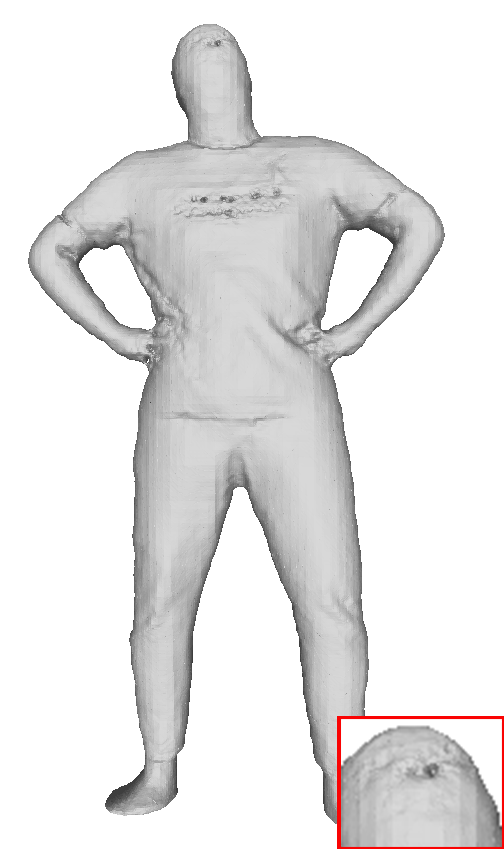}& \includegraphics[width=4in]{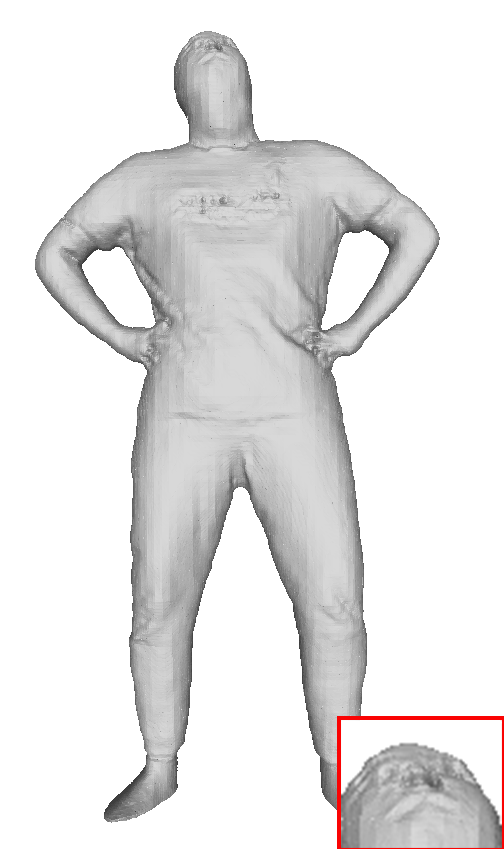} & \includegraphics[width=4in]{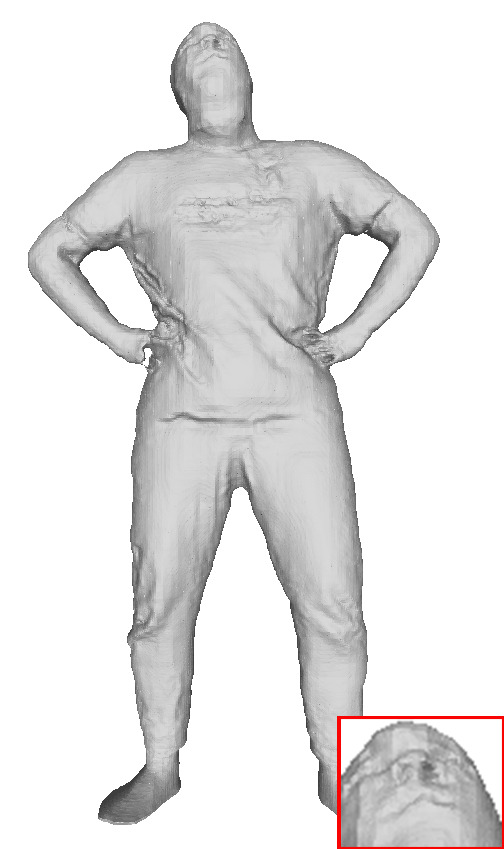}     \\   
\midrule
\includegraphics[width=3.8in]{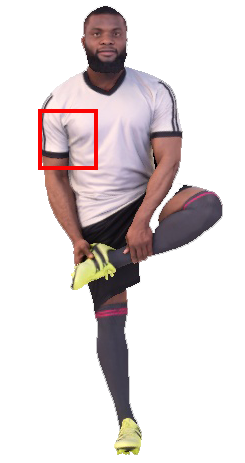}  & \includegraphics[width=4in]{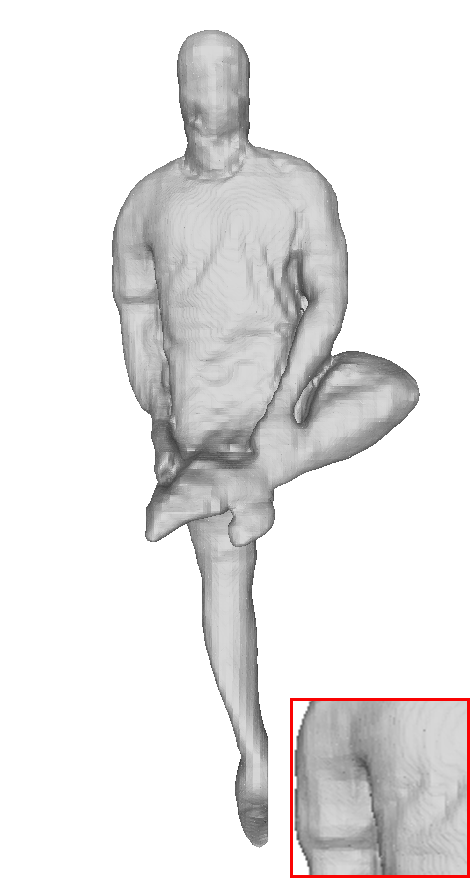} & \includegraphics[width=3.95in]{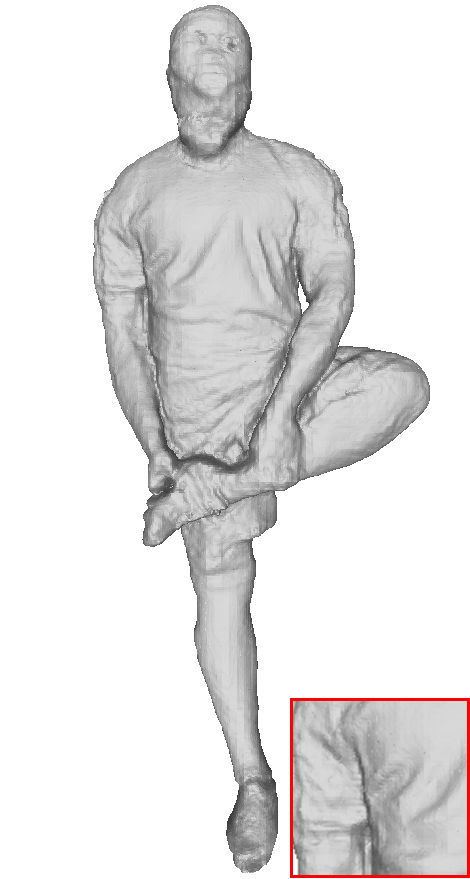} & \includegraphics[width=4in]{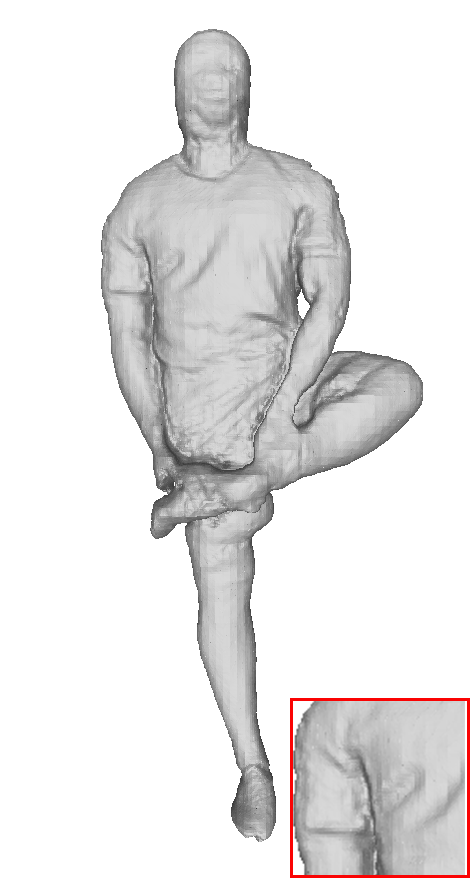}& \includegraphics[width=4in]{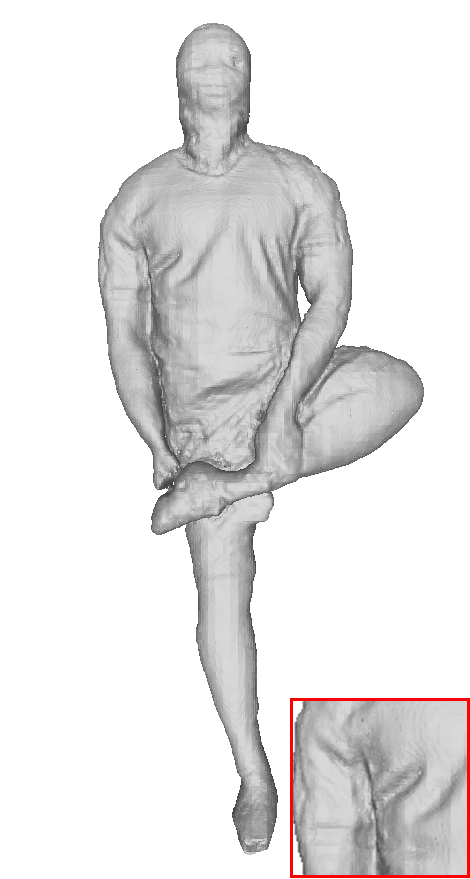} & \includegraphics[width=4in]{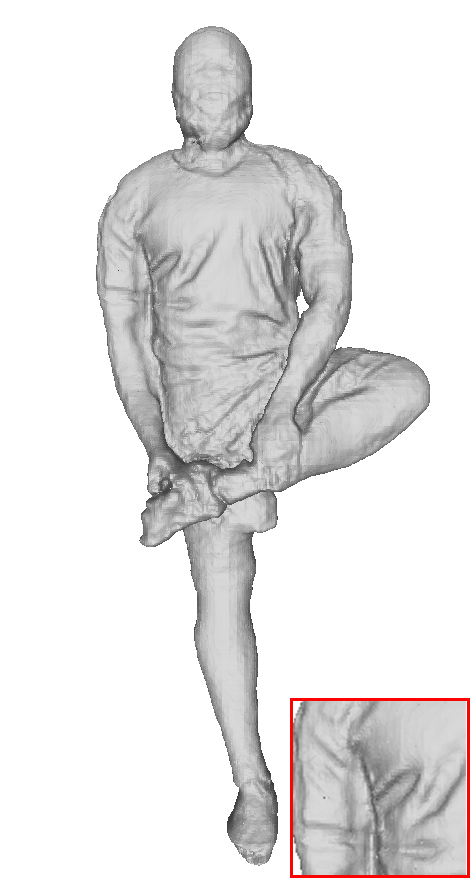} 
\end{tabular}
\end{Huge}}
 \caption{Visual results obtained by changing the architecture of our approach. The upper model is from THuman2.0 while the below is from 3DPeople.}
 \label{fig:loss}
\end{figure*}
Both the quantitative (Table~\ref{tbl:loss}) and the qualitative (Fig.~\ref{fig:loss}) evaluations show the importance of implementing all the different modules of SuRS. If the HR features are not extracted (W/o U-Net), the fine details are lost in the reconstructed shape, which is just a coarse representation of the input image. Less sharp details are obtained when only MR-MLP is applied, with a lower resolution of the final reconstruction. This configuration does not use either the $f_{SR}^{gt}$ ground-truth or the $L_{diff}$ and this deteriorates the performance. In this case, the network does not learn the map from $S_{LR}$ to $S_{HR}$ and the lost information is not retrieved. Similarly, this map is not learned when only SR-MLP is applied and significant lower resolution shapes are obtained compared to when the map is learned (combination of SR-MLP and MR-MLP with HR features), confirming the efficiency of SuRS. The difference between training SuRS with or without $L_{diff}$ is significant proving its efficiency: the wrinkles of the dress are sharper and human body parts are more realistic when $L_{diff}$ is minimised.
%explain the fact of sr when is trained alone and not with the mlp
\subsection{Comparisons}
\label{sec:comp}
\begin{figure}[t!]
\centering
\resizebox{1\linewidth}{!}{\begin{tiny}

\begin{tabular}{ccccc}
HR image             & LR patch &PiFu\cite{saito2019pifu}&$\mathrm{PIFuHD_{no}}$\cite{saito2020pifuhd}             & SuRS (ours) \\
\raisebox{0.025\height}{\includegraphics[width=1in]{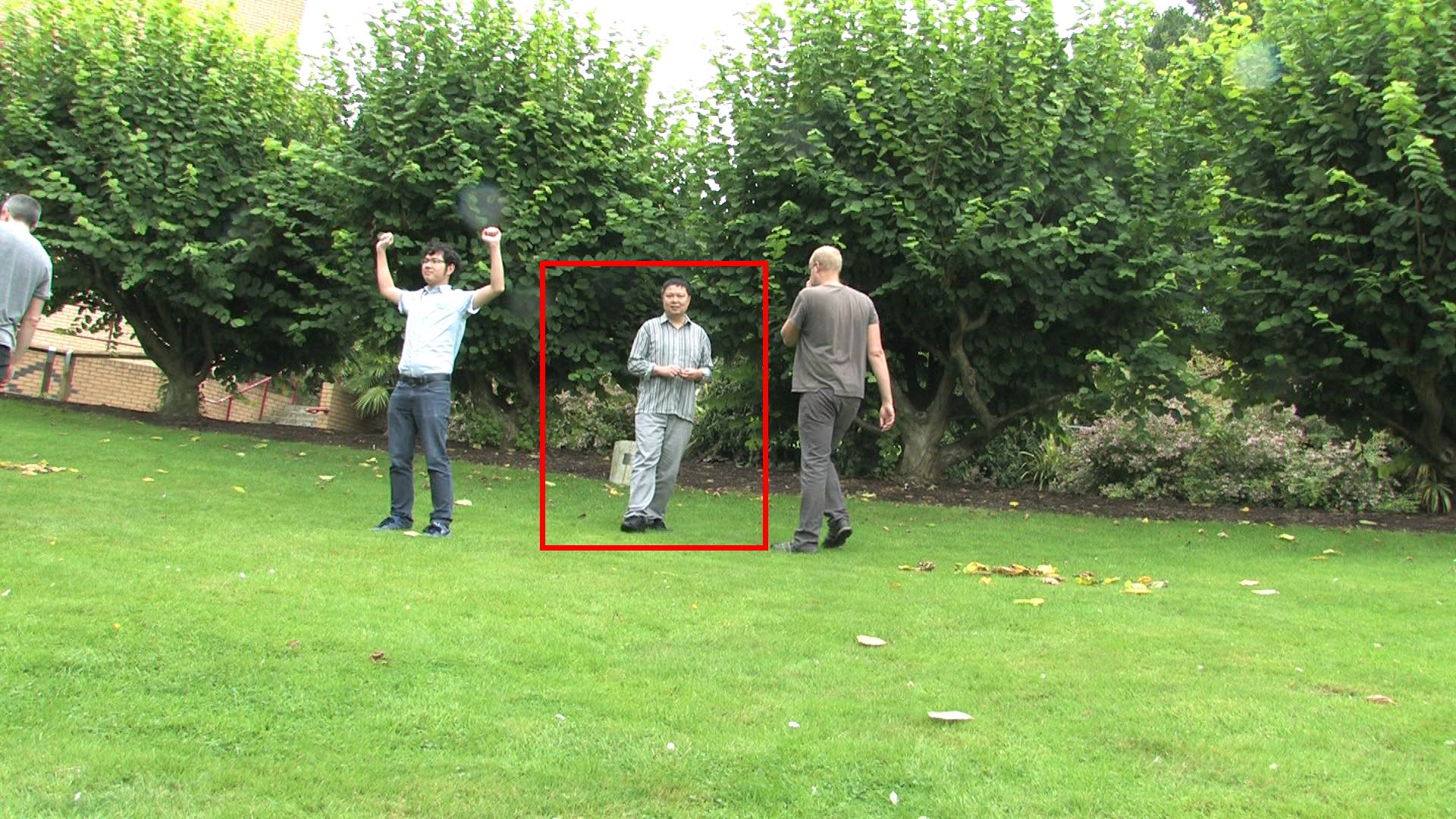} } & \includegraphics[width=0.23in]{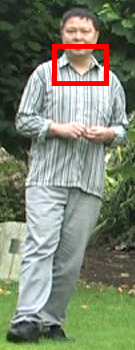} & \includegraphics[width=0.35in]{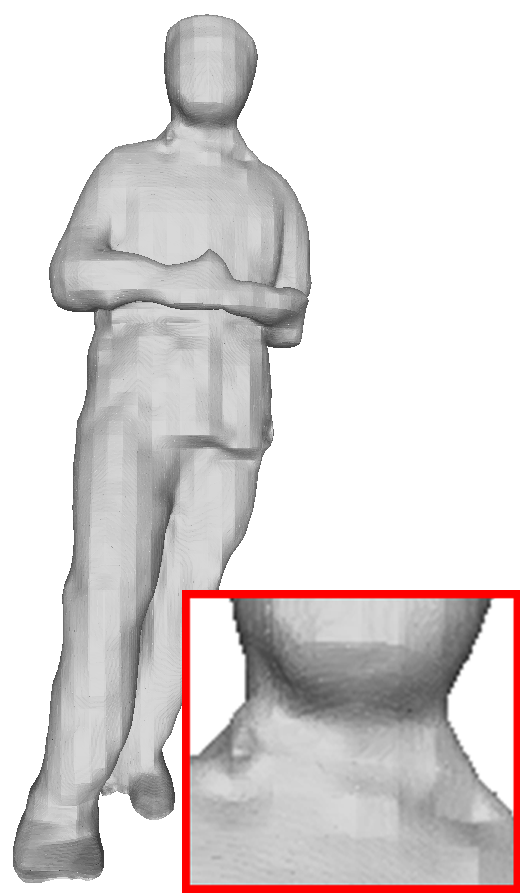} & \includegraphics[width=0.375in]{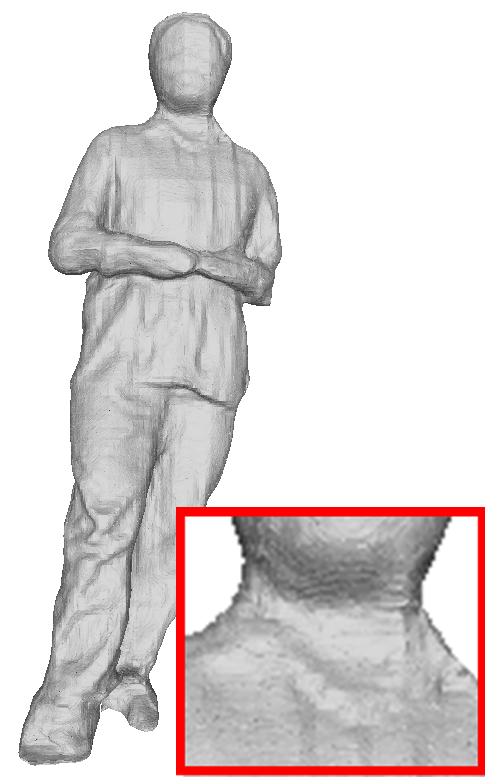} & \includegraphics[width=0.35in]{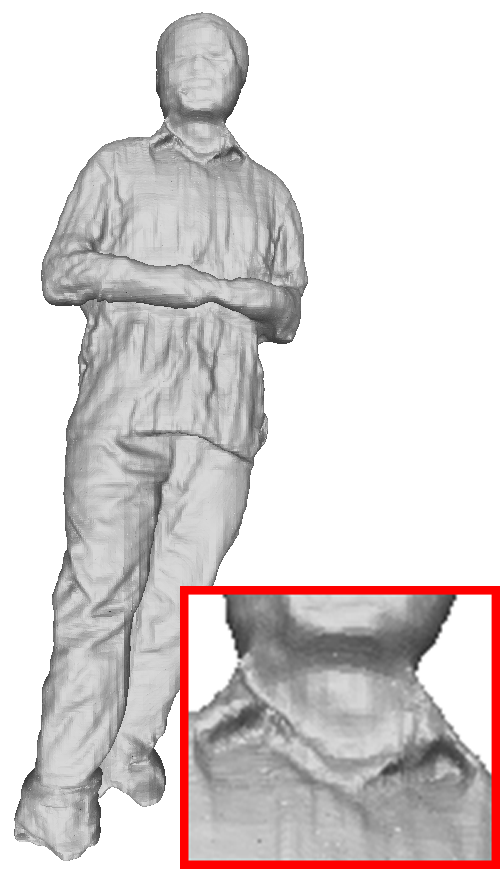}     \\   
\raisebox{0.02\height}{\includegraphics[width=0.92in]{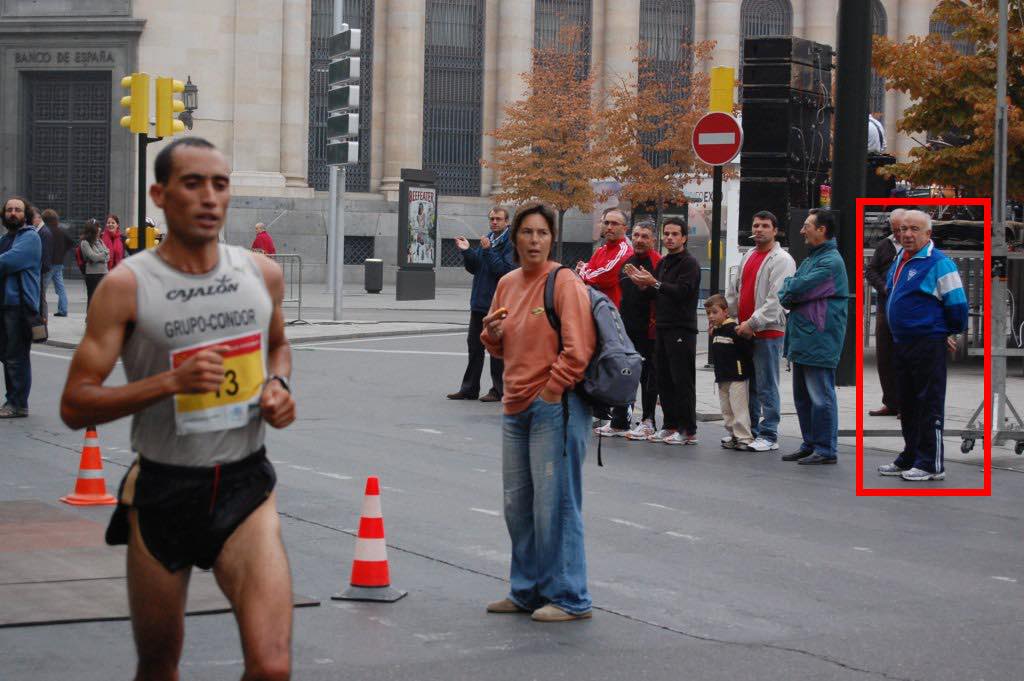}} & \includegraphics[width=0.23in]{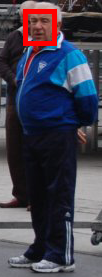} & \includegraphics[width=0.34in]{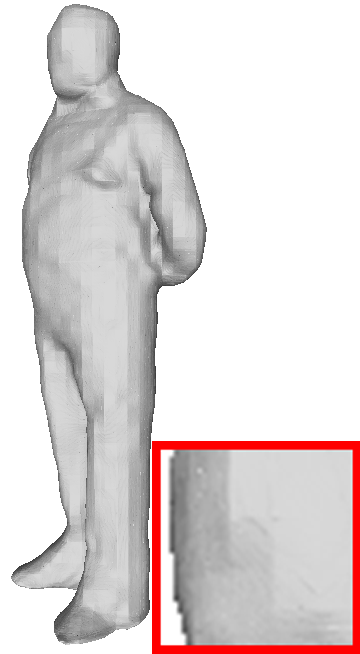} & \includegraphics[width=0.34in]{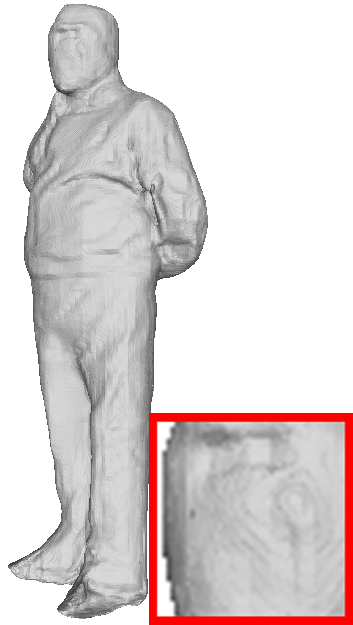} & \includegraphics[width=0.34in]{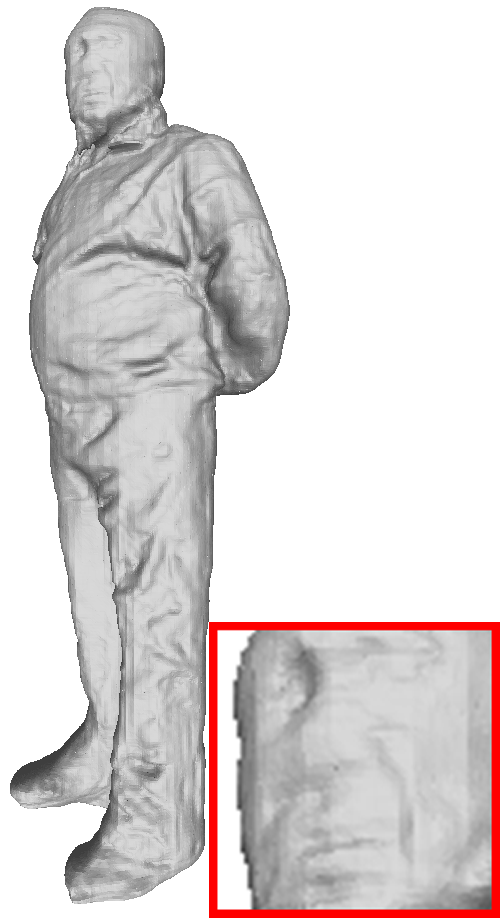} \\
\includegraphics[width=0.87in]{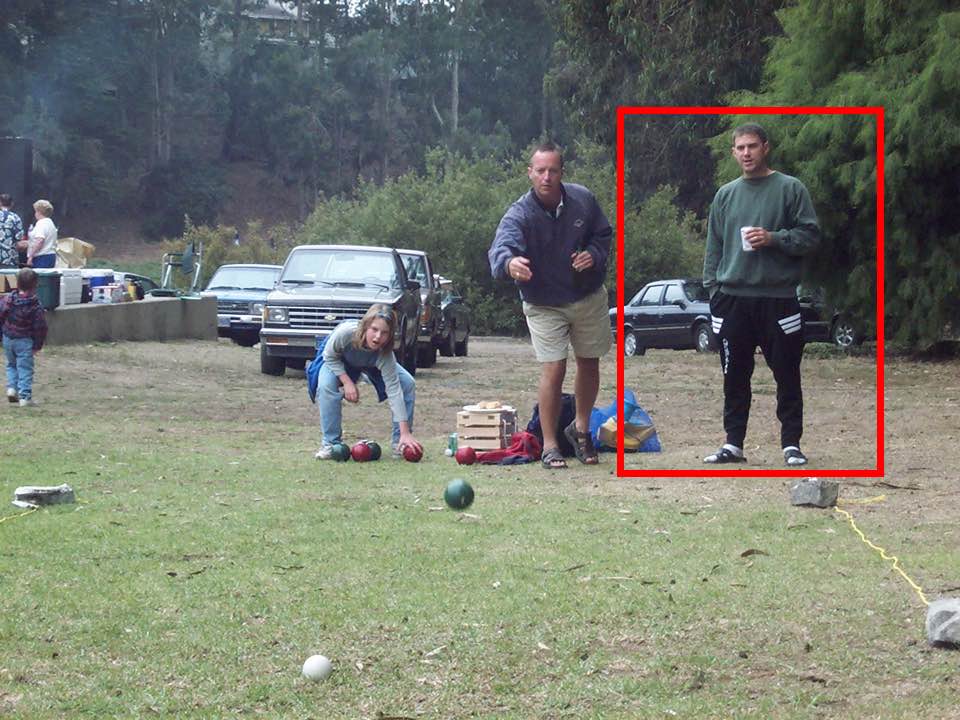}  & \includegraphics[width=0.25in]{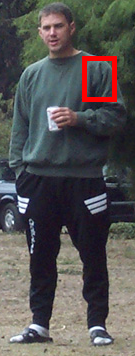} & \includegraphics[width=0.38in]{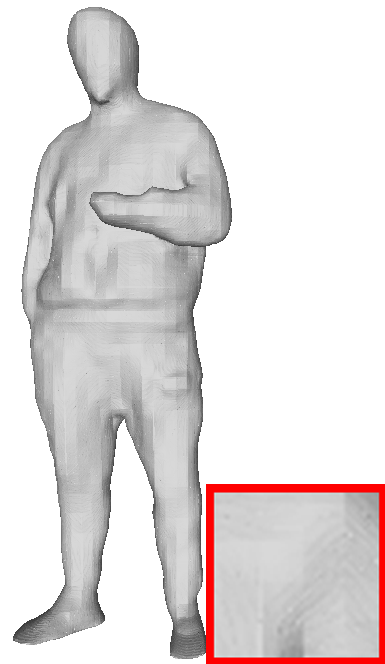} & \includegraphics[width=0.38in]{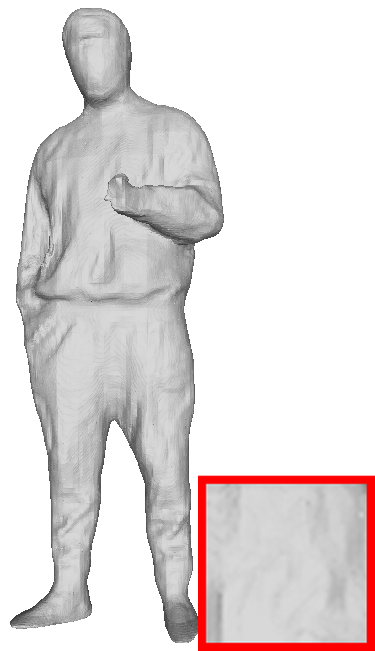} & \includegraphics[width=0.378in]{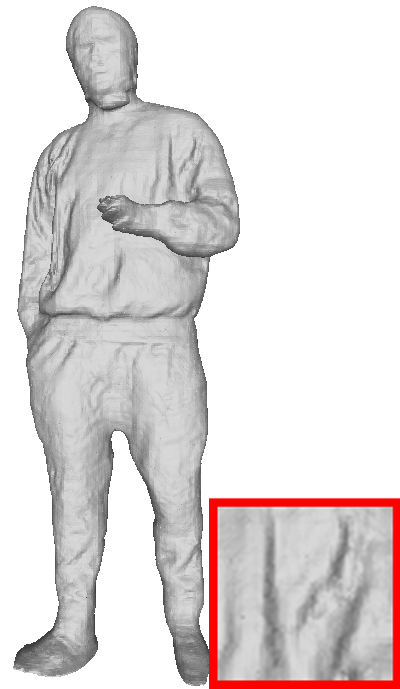} \\
\includegraphics[width=0.845in]{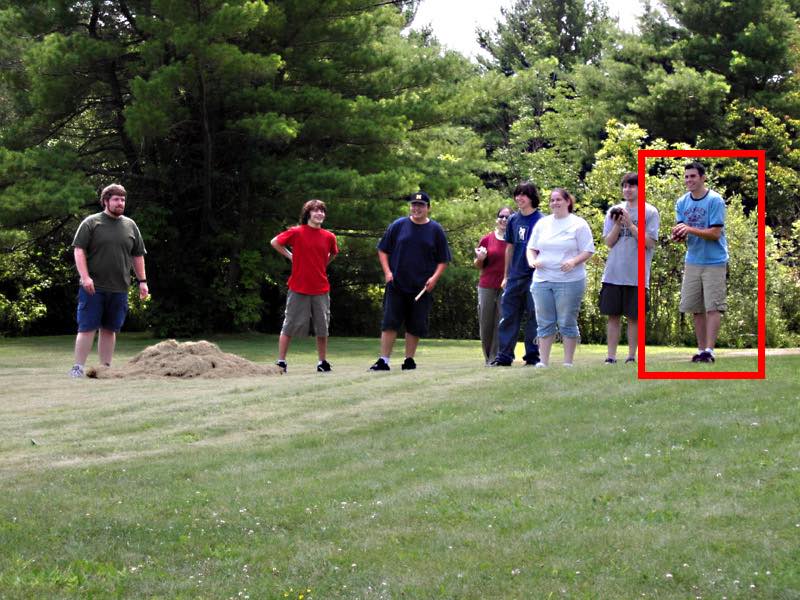}  & \includegraphics[width=0.25in]{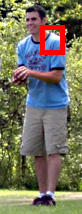} & \includegraphics[width=0.35in]{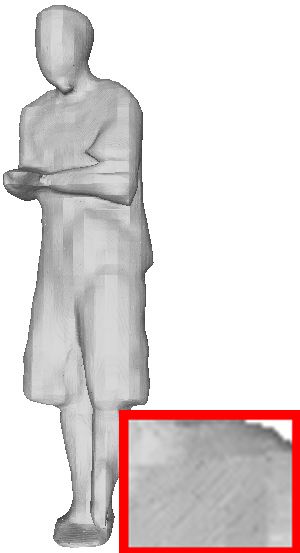} & \includegraphics[width=0.35in]{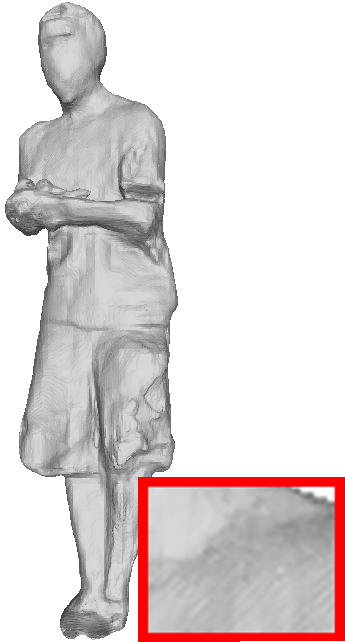} & \includegraphics[width=0.34in]{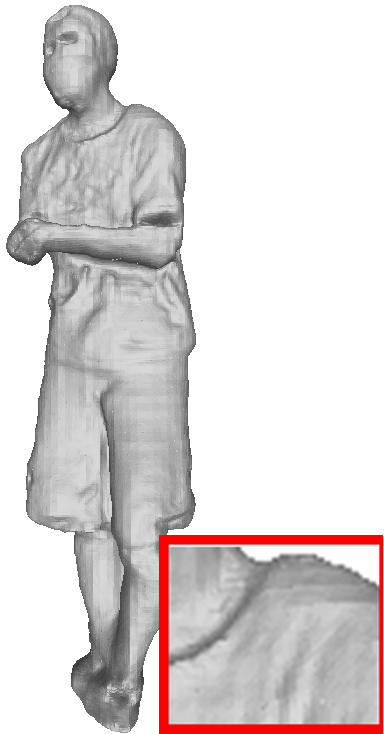} \\
\includegraphics[width=0.745in]{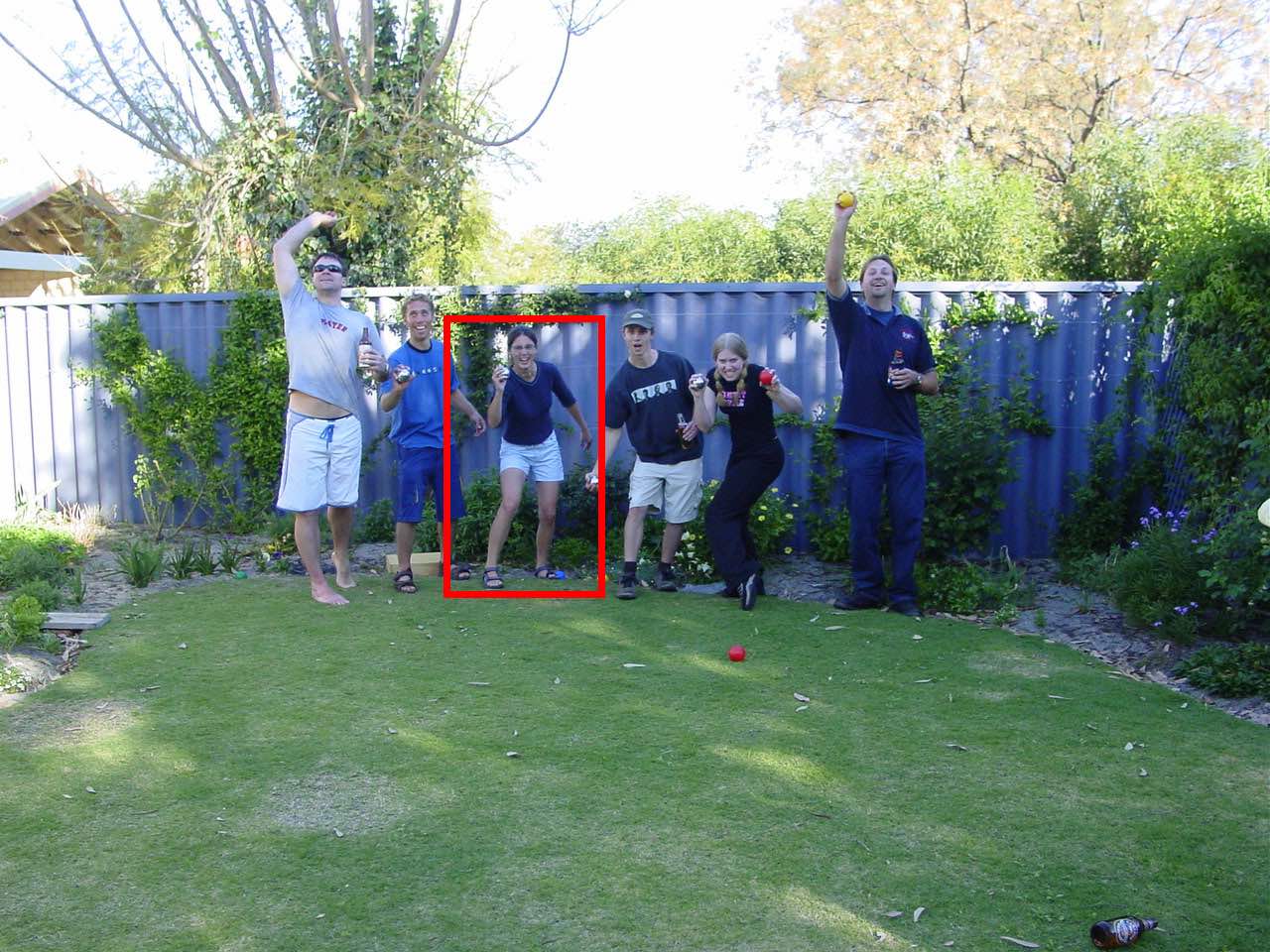}  & \includegraphics[width=0.25in]{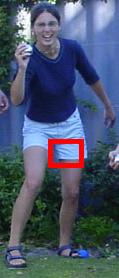} & \includegraphics[width=0.375in]{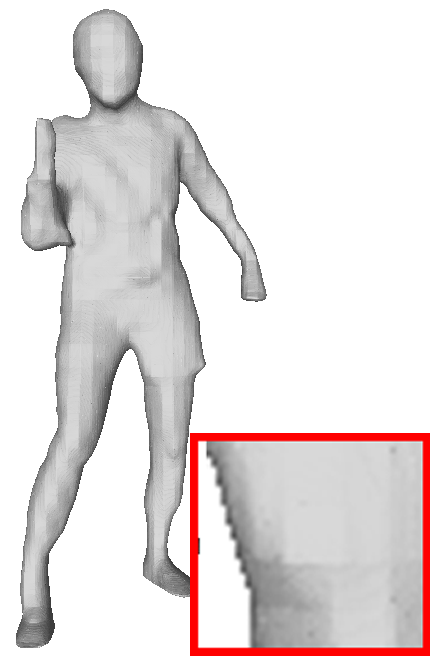} & \includegraphics[width=0.35in]{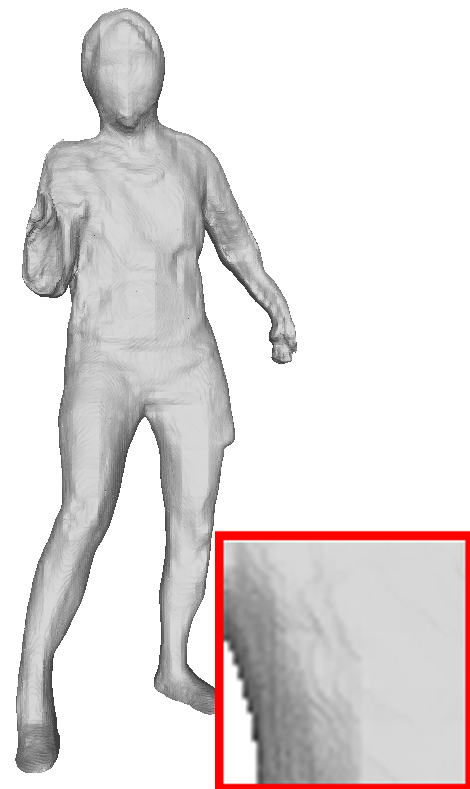} & \includegraphics[width=0.35in]{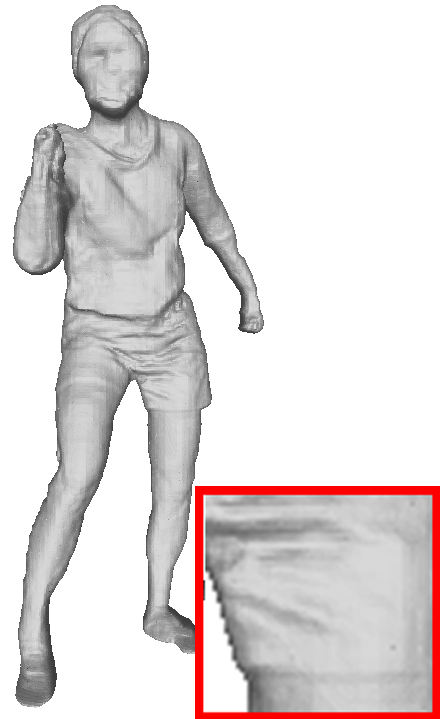} \\
\includegraphics[width=0.755in]{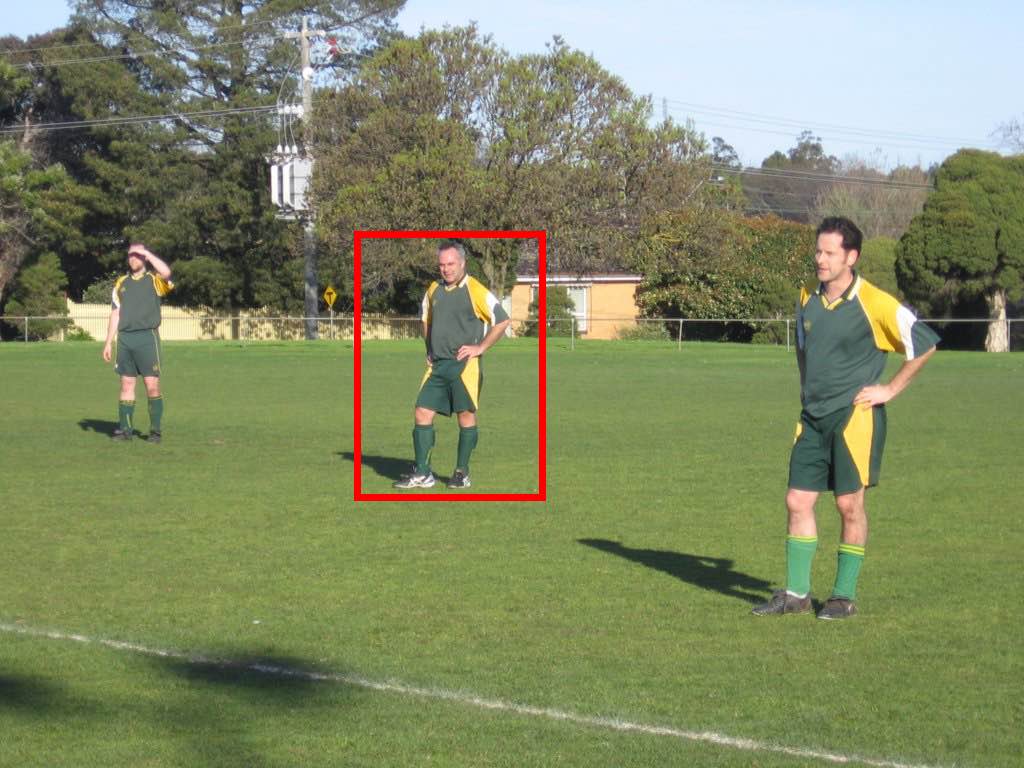} & \includegraphics[width=0.29in]{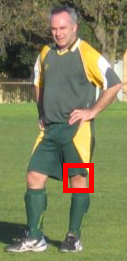} & \includegraphics[width=0.39in]{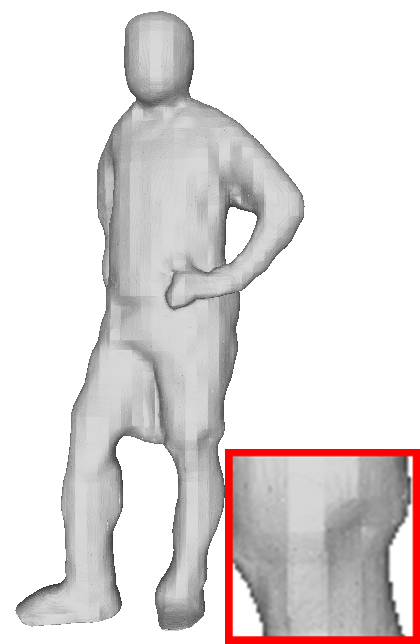} & \includegraphics[width=0.38in]{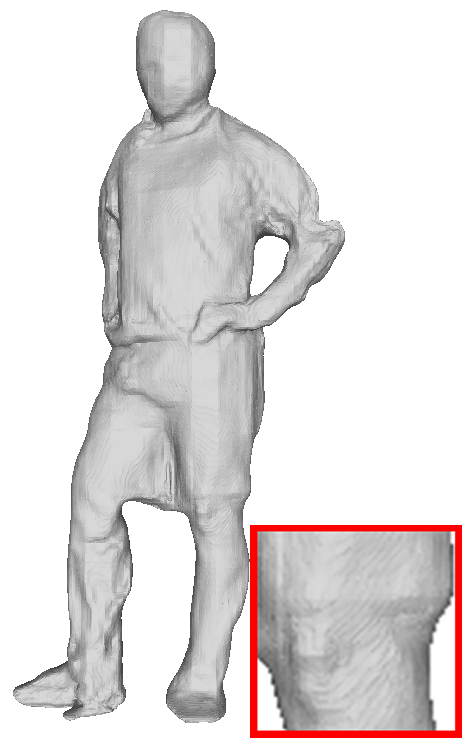} & \includegraphics[width=0.375in]{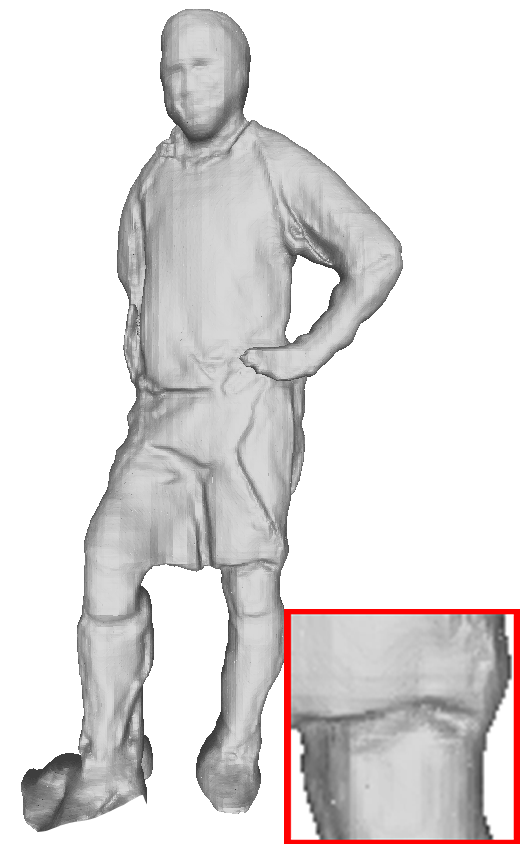} 
\\
\includegraphics[width=0.945in]{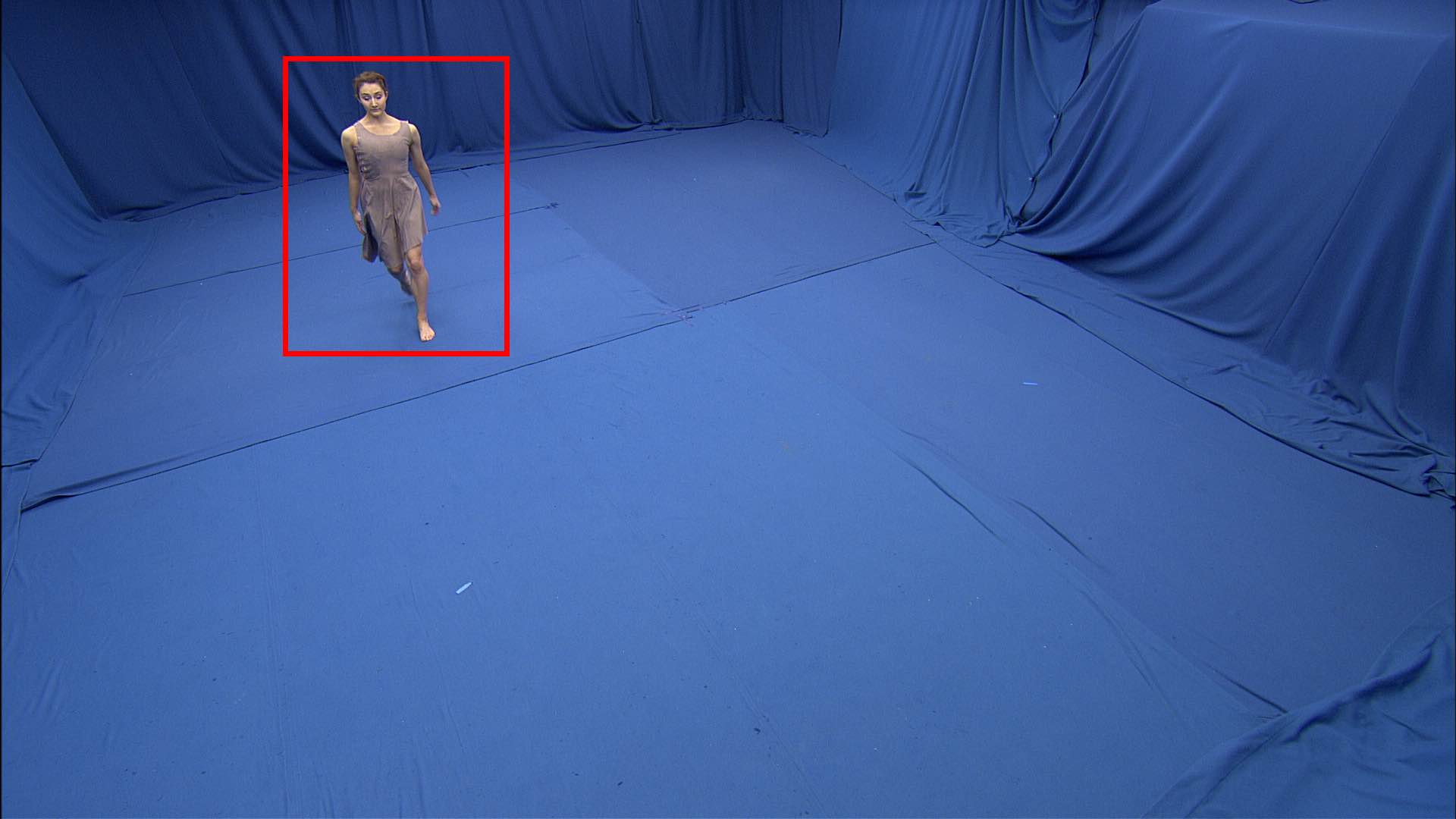} & \includegraphics[width=0.25in]{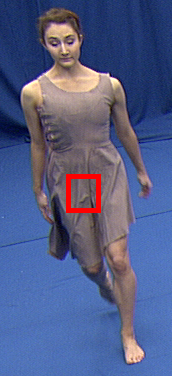} & \includegraphics[width=0.345in]{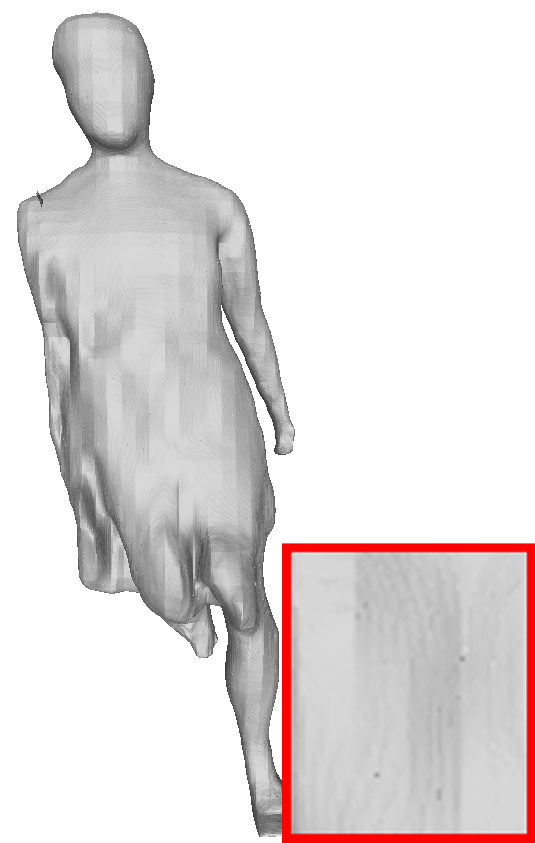} & \includegraphics[width=0.355in]{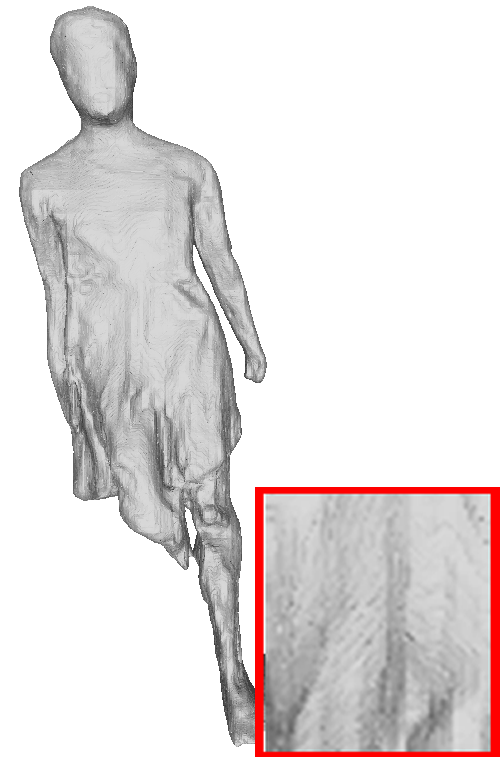} & \includegraphics[width=0.35in]{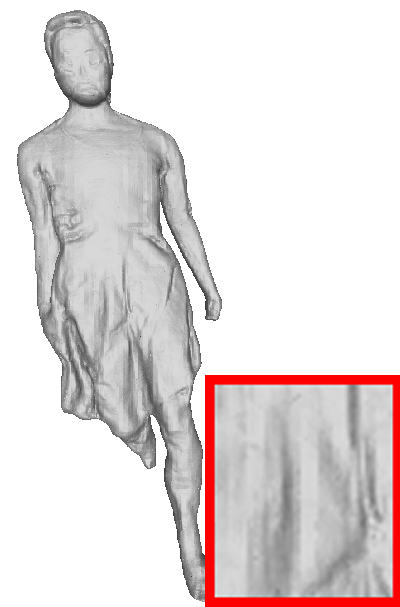} 
\end{tabular}
\end{tiny}}
 %\vspace{-1.1em}
 \caption{Human shapes reconstructed from LR patches of HR images of people.}
 %\vspace{-1.8em}
 \label{fig:comp_HR}
\end{figure}

\begin{table}[t!]
    \centering
    {
    \caption{Quantitative comparisons between state-of-the-art approaches with LR input image for training and testing. The highest scores are highlighted in red while the second highest scores are blue.}
\resizebox{0.5\textwidth}{!}{\begin{tabular}{|c|l|lll|lll|}
\hline
\multirow{2}{*}{\bm{$N_I$}}                                                 & \multirow{2}{*}{\textbf{Methods}} & \multicolumn{3}{c|}{\textbf{THuman2.0}}           & \multicolumn{3}{c|}{\textbf{3D people}}           \\ \cline{3-8} 
                                                                       &                                   & CD             & Normal          & P2S            & CD             & Normal          & P2S            \\ \hline
\multirow{6}{*}{\begin{tabular}[c]{@{}c@{}}2\\ 5\\ 6\end{tabular}}     & DeepHuman~\cite{zheng2019deephuman}                         & 1.956          & 0.1465          & 2.063          & 1.703          & 0.1253          & 1.733          \\
                                                                       & PIFu~\cite{saito2019pifu}                              & 1.518          & 0.1218          & 1.647          & 1.519          & {\color[HTML]{3531FF}0.1198}         & 1.581          \\
                                                                       & $\mathrm{PIFuHD_{no}}$                        & 1.308          & 0.1237          & 1.346          & 1.440          & 0.1595          & 1.464          \\
                                                                       & PIFuHD~\cite{saito2020pifuhd}                             &{\color[HTML]{3531FF}1.032}         &{\color[HTML]{3531FF}0.1116}          &{\color[HTML]{FF0000}1.046} &{\color[HTML]{3531FF}1.062}          & 0.1257          &{\color[HTML]{FF0000}1.037} \\
                                                                       & PaMIR~\cite{zheng2021pamir}                              & 1.713          & 0.1341          & 1.818          & 1.644          & 0.1410          & 1.740          \\
                                                                       & Geo-PIFu~\cite{he2020geo}                               & 1.786          & 0.1473         & 1.724          & 1.853          & 0.1658          & 1.652          \\
                                                                       & SuRS (ours)                      &{\color[HTML]{FF0000}0.931} &{\color[HTML]{FF0000}0.1065} &{\color[HTML]{3531FF}1.151}         &{\color[HTML]{FF0000}1.057} &{\color[HTML]{FF0000}0.1127} &{\color[HTML]{3531FF}1.247}         \\ \hline
\end{tabular}

\color[HTML]{FF0000}
\color[HTML]{3531FF}                                                                      }
%\vspace{-0.8em}
%\vspace{-2.em}
\label{tbl:quant_LR}
}
% \hfill
% \parbox{.48\linewidth}{
% \caption{Quantitative comparisons between state-of-the-art approaches with higher sizes ($N_{I}=512,1024$) input image for training and testing.}
% \resizebox{0.45\textwidth}{!}{\input{table/comparison_HR}}
% %\vspace{-0.8em}
% %\vspace{-2.em}
% \label{tbl:quant_HR}
% }
\end{table}
\begin{figure}[t!]
\centering
\resizebox{0.85\linewidth}{!}{\begin{tabular}{cccccccccccc}
$N_{I}$                                               &       Input image & DeepHuman~\cite{zheng2019deephuman} & \multicolumn{2}{c}{PIFu~\cite{saito2019pifu}} & $\mathrm{PIFuHD_{no}}$ & \multicolumn{2}{c}{PIFuHD~\cite{saito2020pifuhd}} & PaMIR~\cite{zheng2021pamir}       & \multicolumn{2}{c}{Geo-PIFu~\cite{he2020geo}}  & SuRS (ours) \\
\raisebox{6\height}{\begin{tabular}[c]{@{}c@{}}2\\ 5\\ 6\end{tabular}}     & {\includegraphics[width=1.5in]{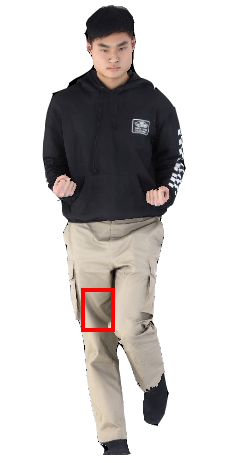}} & \includegraphics[width=0.9in]{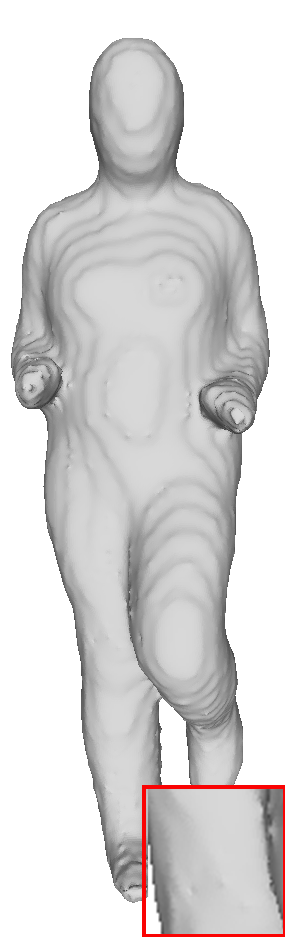}         & \multicolumn{2}{c}{\includegraphics[width=0.9in]{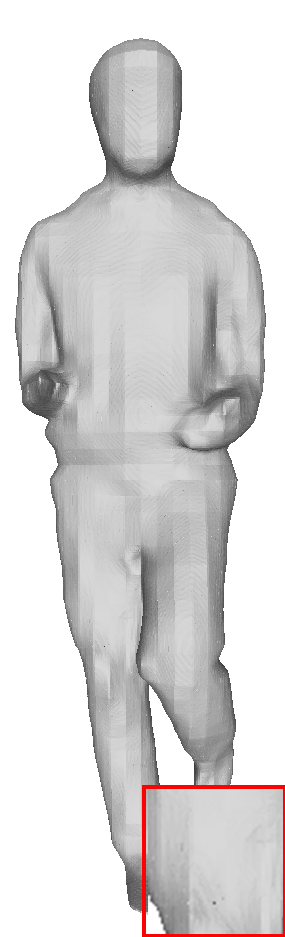}}    & \includegraphics[width=0.9in]{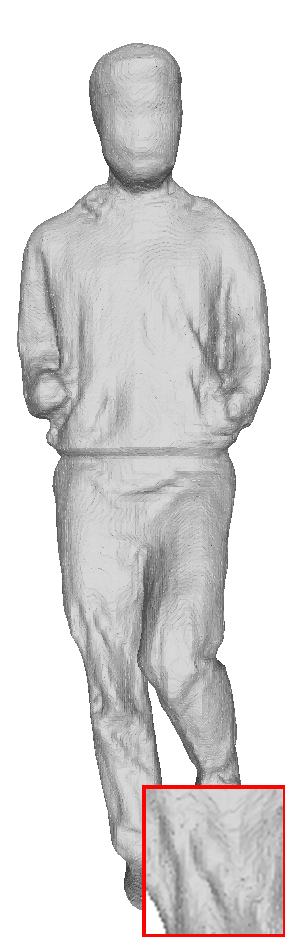}         & \multicolumn{2}{c}{\includegraphics[width=0.9in]{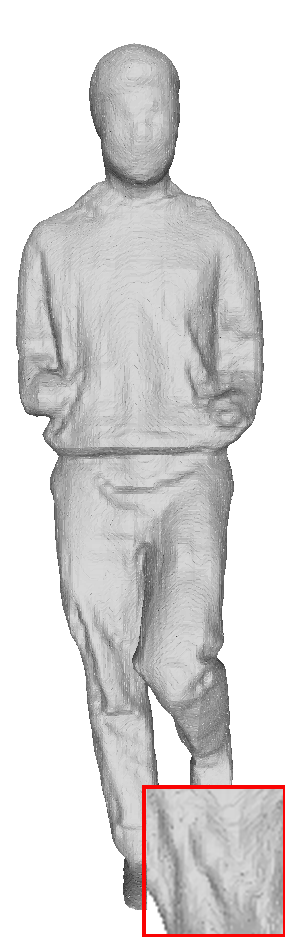}}       & \includegraphics[width=0.9in]{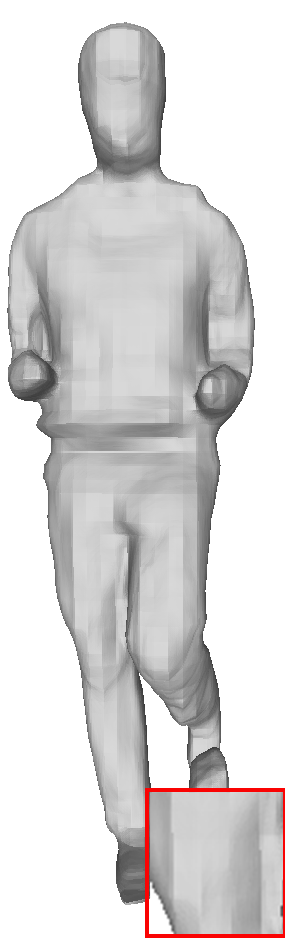}        & \multicolumn{2}{c}{\includegraphics[width=0.98in]{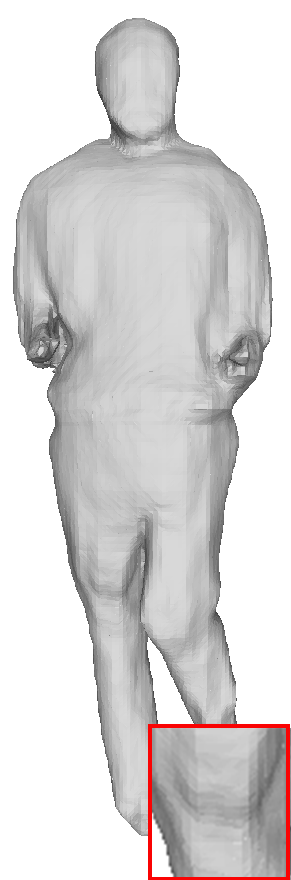}} & \includegraphics[width=0.9in]{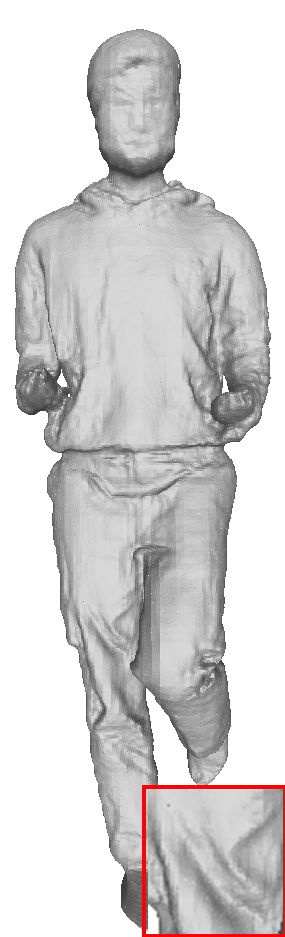} 
\\
\midrule
\\
\raisebox{5\height}{\begin{tabular}{c} 2\\ 5\\ 6\end{tabular} }    & {\includegraphics[width=1.35in]{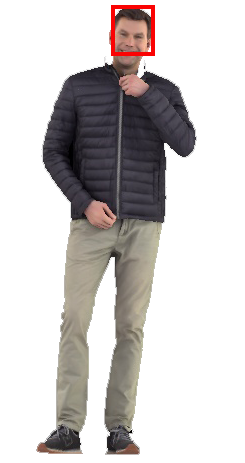}} & \includegraphics[width=1in]{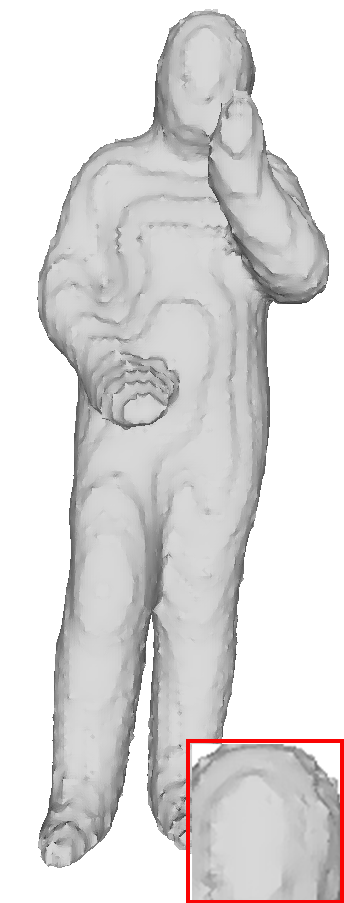}         & \multicolumn{2}{c}{\includegraphics[width=1.03in]{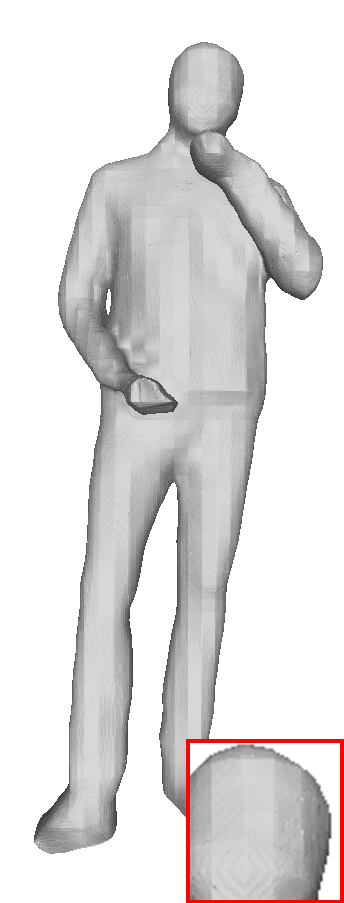}}    & \includegraphics[width=1in]{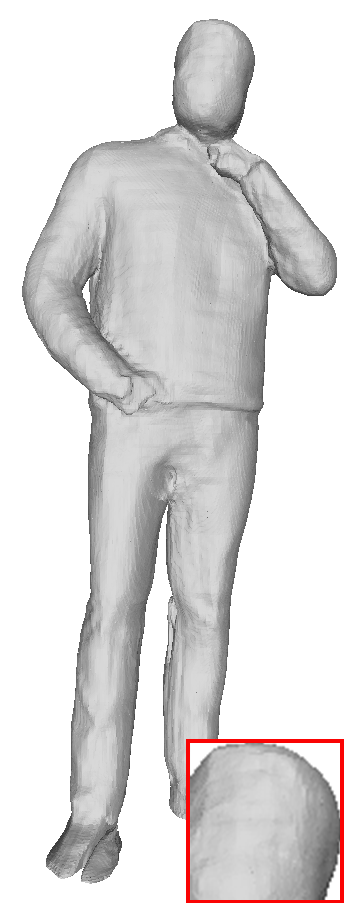}         & \multicolumn{2}{c}{\includegraphics[width=1in]{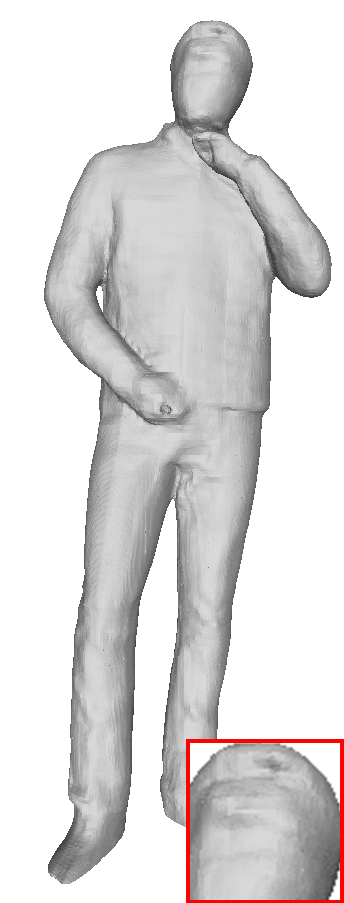}}       & \includegraphics[width=0.97in]{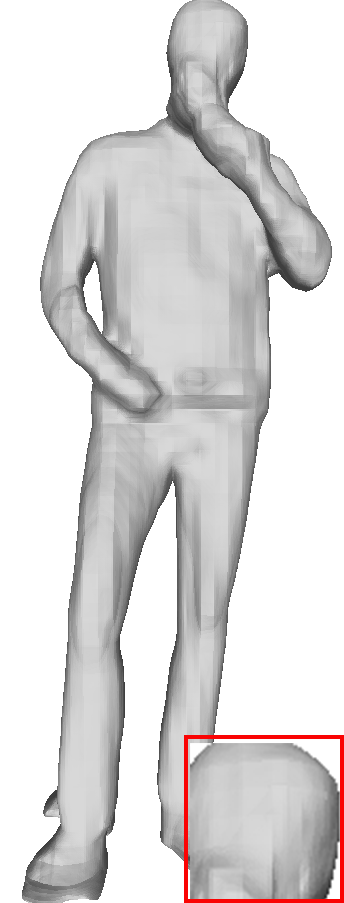}        & \multicolumn{2}{c}{\includegraphics[width=0.98in]{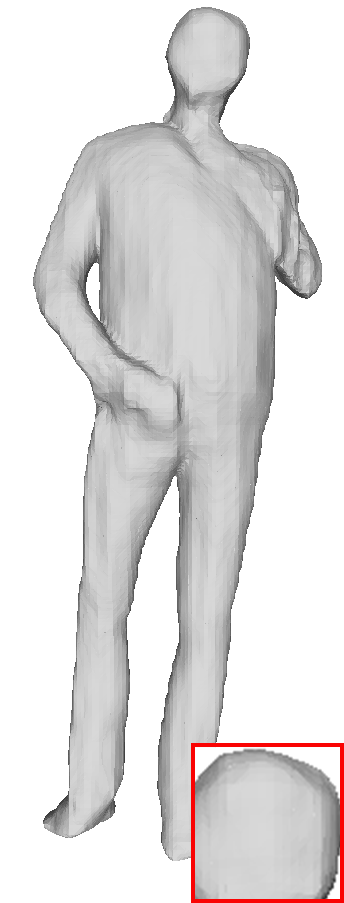}}   & \includegraphics[width=1in]{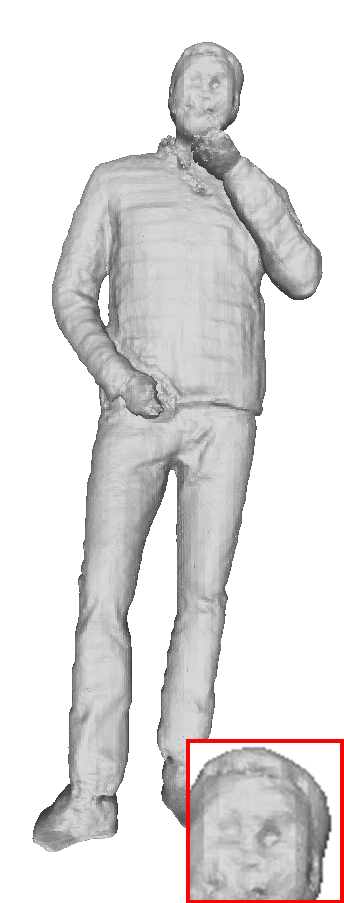}             \\
\end{tabular}}
 %\vspace{-1.1em}
 \caption{Visual comparisons using LR input image for training and testing. The upper model is from THuman2.0, the below one is from 3DPeople.}
 %\vspace{-1.8em}
 \label{fig:comp_LR}
\end{figure}
\noindent We compare SuRS with related works on 3D human digitization from a single image. We evaluate DeepHuman~\cite{zheng2019deephuman}, which do not use implicit representation but leverage parametric models to facilitate the reconstruction. We then compare SuRS with approaches that represent the 3D surfaces with implicit representation. Among these, PIFu~\cite{saito2019pifu} is the only one which uses only single RGB images. PIFuHD~\cite{saito2020pifuhd} exploits front and back normal maps while PaMIR~\cite{zheng2021pamir} and Geo-PIFu~\cite{he2020geo} leverage parametric models. We train and test PIFuHD without using normal maps ($\mathrm{PIFuHD_{no}}$). These approaches are trained and tested with the same datasets as SuRS except for Geo-PIFu, which cannot pre-processed THuman2.0 shapes. See supplementary material for further comparisons.%Since the shapes of THuman2.0 cannot be pre-processed by Geo-PIFu, we use its pre-trained weights to test. 
\\\textbf{Qualitative evaluation: }Fig.~\ref{fig:comp_LR} illustrates two shapes reconstructed from a LR input image of sizes $N_{I}=256$. Compared to other methods, SuRS reconstructs significantly higher resolution shapes, which contain the highest level of fine details in clothes and faces. Related works produce coarse shapes with smooth details even if other data such as normal maps or parametric models is leveraged.
\\\textbf{Quantitative evaluation:} Our approach outperforms related works when only LR input images are used in both training and testing. SuRS achieves the highest values of the considered metrics, proving its superiority over all the considered approaches (Table~\ref{tbl:quant_LR}). SuRS outperforms also all the other approaches that use auxiliary data for the CD and Normal metrics. PiFUHD achieves lower values of P2S due to its use of back normal maps, which improve its performance on parts of the model that are not seen in just a single image.
\\\textbf{Real data:} We qualitatively evaluate our approach on images of people captured in real scenarios. More specifically, we reconstruct the 3D shape of a person from an HR image ($> 1024$) where the person resolution is at a relatively low-resolution sub-image of the original HR image ($< 350 \times 350$). We select random images from various datasets~\cite{li2007and,Johnson10} and we compare SuRS with only the approaches that do not use auxiliary data in the reconstruction, namely PiFU and $\mathrm{PIFuHD_{no}}$. For the reconstruction, a LR human body patch is cropped from the HR image. SuRS significantly outperforms the other approaches, reproducing higher level of fine details on the reconstructed shapes, which are coarser with smoother details if reconstructed by the other approaches (Fig.~\ref{fig:comp_HR}).
\\\textbf{Limitations:} our method cannot super-resolve parts of the human body that are not visible on the input image: since no auxiliary data are provided, SuRS reconstructs a coarse geometry of the hidden parts of the body. Like existing works, it may generate incorrect body structures when the input model presents features that significantly differ from the ones seen during training. It may also suffer problems related to depth ambiguity. Examples in supplementary material.
%We do not train SuRIF dueto GPU memory limitations (to process features of 2048 size). 
% \begin{table}[t!]
%     \centering
%     \caption{Quantitative comparisons between state-of-the-art approaches with different sizes of the input image for training and testing. The highest scores are highlighted in red while the second highest scores are blue for each size of input image. The bold red figures are the highest across all categories. The `no' subscript means that normal maps are not used. The `up' superscript means that the input image is upscaled from $256$ to $1024$.}
% \resizebox{0.45\textwidth}{!}{\input{table/comparisons}}
% %\vspace{-0.8em}
% %\vspace{-2.em}
% \label{tbl:quant}
% \end{table}
\section{Conclusion}
% \begin{table*}[!]
% \centering
% \resizebox{0.95\linewidth}{!}{\input{image_files/comparison_3}}
%  %\vspace{-1.1em}
%  \captionof{figure}{Visual comparisons using different sizes of the input image for training and testing. The upper model is from THuman2.0, the below one is from 3DPeople. The `no' subscript means that normal maps are not used. The `up' superscript means that the input image is upscaled from $256$ to $1024$.}
%  %\vspace{-1.8em}
%  \label{fig:comp}
% \end{table*}
\noindent We propose Super-Resolution shape represented by an high-detail implicit representation and we tackle the problem of 3D human digitization from a single low-resolution image. To achieve this, we propose a novel architecture with a customised novel loss that can reconstruct super-resolution shape from a low-resolution image. As demonstrated by the evaluation, the reconstructed surfaces contain significantly higher level of details compared to the outputs of related methods when low-resolution images are used. The resolution of the shape obtained from a low-resolution image is significantly higher than the one obtained by state-of-the-art works that leverage also auxiliary data such as normal maps and parametric models. As future works, a super-resolution texture representation will be investigated and the approach will be extended to dynamic shape.
\\\textbf{Acknowledgement}: this research was supported by UKRI EPSRC Platform Grant EP/P022529/1
%%%%%%%%% REFERENCES
\bibliographystyle{splncs04}
\bibliography{eccv_latest}
\end{document}